%% file: Manuscript.tex
\documentclass[preprint,12pt,authoryear]{elsarticle}
\usepackage[T1]{fontenc}
\usepackage[utf8]{inputenc}
\usepackage{lmodern}
\usepackage[a4paper,margin=2.5cm]{geometry}
\usepackage{graphicx}
\usepackage{booktabs}
\usepackage[font=small]{caption}
\usepackage{placeins}
\usepackage{float}
\usepackage{amsmath,amssymb}
\usepackage{xcolor}
\usepackage{hyperref}
\definecolor{ManuscriptLink}{HTML}{155CA2}
\hypersetup{colorlinks=true,linkcolor=ManuscriptLink,citecolor=ManuscriptLink,urlcolor=ManuscriptLink,
pdftitle={From syllable representation via gesture generation to articulation: A spiking neural model},pdfauthor={Bernd J. Kröger}}
\biboptions{round,semicolon}
\bibpunct{(}{)}{;}{a}{,}{,}
\makeatletter
\DeclareRobustCommand{\citeand}{\ifNAT@swa\&\else and\fi}
\def\ps@pprintTitle{\let\@oddhead\@empty\let\@evenhead\@empty\def\@oddfoot{\hfil\thepage\hfil}\let\@evenfoot\@oddfoot}
\makeatother
\newcommand{\norm}[1]{\left\lVert#1\right\rVert}
\newcommand{\code}[1]{\texttt{#1}}
\begin{document}
\clubpenalty=3000
\widowpenalty=3000
\emergencystretch=1em
\begin{frontmatter}
\title{From syllable representation via gesture generation to articulation: A spiking neural model\tnoteref{preprint}}
\tnotetext[preprint]{Author preprint, version 2, 27 November 2026. This preprint is distributed under the Creative Commons Attribution 4.0 International license (CC BY 4.0).}
\author[aff1]{Bernd J.~Kröger\corref{cor1}}
\address[aff1]{Faculty of Medicine, RWTH Aachen University, Pauwelsstra\ss{}e 30, 52074 Aachen, Germany}
\cortext[cor1]{Corresponding author.}
\ead{bernd.kroeger@rwth-aachen.de}
\begin{abstract}
\input{abstract}
\end{abstract}
\end{frontmatter}
\input{01-introduction}
\input{02-representation}
\input{03-neural-timing}
\input{04-articulation}
\input{task-realization}
\input{simulation-corpus}
\input{05-simulations}
\input{06-discussion}

\input{07-resources}
\input{08-conclusion}
\input{declarations}
\begingroup
\raggedright

\input{references}
\endgroup
\clearpage
\appendix
\setcounter{table}{0}
\setcounter{figure}{0}
\setcounter{equation}{0}
\renewcommand{\thetable}{\Alph{section}.\arabic{table}}
\renewcommand{\thefigure}{\Alph{section}.\arabic{figure}}
\renewcommand{\theequation}{\Alph{section}.\arabic{equation}}
\input{appendix}
\end{document}

%% file: abstract.tex
Speech production requires a transformation from phonological representations into coordinated vocal-tract movements. This paper presents a spiking neural sensorimotor model within the Neural Engineering Framework and Semantic Pointer Architecture. Each syllable occurrence has a neural oscillator whose evolving phase coordinates its gestures and initiates the next syllable. Phase-dependent commands recruit separate neural populations for vocalic, oral-constriction, velopharyngeal, and glottal actions (gestures). Neural ramp integrators shape the rise and fall of gesture activation. Together with articulatory targets, these trajectories determine the extent and time course of articulatory movements. Opponent neural populations provide control signals to a geometric articulatory model.

Thirty reference recordings cover an input corpus of eight lexical words, an alternative realization of one word, a syllable sequence, and isolated syllables. Two additional paired simulation subcorpora of sixty runs each, covering two words and one syllable sequence, assess activation reliability across neural realizations. Adjusting gesture command amplitudes increased successful realizations from 57/60 to 60/60 under the specified criteria, with ramp rate factors held constant.

The model demonstrates how neural oscillator and ramp dynamics generate articulatory timing under specified phasing rules. It complements a preceding mental-lexicon model through an ordered syllable interface. Continuous online execution across all components and connection to the acoustic extension remain further integration steps. Accessible code and synchronized visualization support inspection of the neural activation trajectories and resulting vocal-tract movements.

%% file: 01-introduction.tex
\section{Introduction}
Retrieving a word's phonological form provides the linguistic material for speaking it. Articulation additionally requires a temporal organization: the tongue, lips, velum, and glottis must perform partially overlapping actions with appropriate relative timing. A production model must therefore explain how a selected representation becomes an evolving pattern of motor activity. This transition is particularly informative in a neural model because representation, selection, and timing can be expressed within the same computational framework.

The Neural Engineering Framework (NEF) provides methods for representing variables, computing transformations, and implementing dynamics in populations of spiking neurons \citep{eliasmith2003}. The Semantic Pointer Architecture (SPA) uses distributed vector representations to organize higher-level processing \citep{eliasmith2013}. For articulation, an especially useful NEF component is a recurrent population representing a neural oscillator. Its state advances through a cycle and thus supplies a continuously available temporal coordinate. Gestures can be associated with phase regions of this cycle, and a later syllable can begin when the current syllable reaches a specified phase. This is the principal reason for adopting neural oscillators: they provide a neural mechanism for turning relative temporal structure into activity that unfolds in real time.

The articulatory organization follows the central gestural idea that speech involves coordinated, overlapping actions. A gesture specifies an action such as forming a tongue constriction, rounding the lips, or opening the velum, together with its target and temporal organization. Articulatory Phonology makes this organization explicit \citep{browman1992}; task dynamics relates gestural targets to movement generation \citep{saltzman1989}. Each action can have its own time interval: an oral closure can overlap in time with vowel formation, and velum opening can begin before a nasal closure and continue through part of its release. Segment labels remain useful for specifying a syllable program, while gesture-specific timing determines its realization. Here, neural populations generate the time-varying activation of these actions, and an articulatory interface translates their combined effects into movement.

This organization also draws on the broader motor-control concept of coordinative structures: task-specific cooperation among effectors \citep{kelso1986}. Phase relations provide a way to describe coordination and its reorganization in movement \citep{haken1985}, while later gestural planning models use coupled oscillators to organize speech gestures \citep{nam2009}. These ideas motivate making gesture generation an explicit intermediate stage between syllable representation and articulation. In our approach, separate phase commands and recurrent ramps expose when each gesture is recruited and how its activation develops. The present implementation supplies these distinct variables; self-organized phase coupling and compensatory responses to perturbations remain extensions (Section~\ref{sec:coordination}).

\citet{kroeger2023} described syllable sequencing, neural oscillators, and premotor gesture activation in an NEF-SPA speech-production model. The present paper develops the technical account from these components through to explicit articulatory control and visualization. Its focus is the relationship between the objects used in the implementation: a syllable occurrence, its neural oscillator state, a phase-dependent gesture command, the resulting activation trajectory, and the parameters controlling vocal-tract geometry. Making these relationships inspectable is valuable both for interpreting neural traces and for extending the model.

Reliable neural execution requires both successful initiation of the neural syllable oscillators and recruitment of the intended gestures. In his speech-production implementation, \citet[pp.~118--119]{bekolay2016} discusses limitations of brief neural oscillator-start pulses and introduces a target-directed reset. We therefore examine possible activation failures in the present architecture, assessing neural oscillator initiation and gesture recruitment separately across neural realizations and gesture command amplitudes.

The present paper describes the sensorimotor model part and its interface to the articulatory-acoustic model, following syllable representations through neural gesture generation and control to vocal-tract movement. The cognitive model part is described in a companion paper presenting a spiking neural implementation of the mental lexicon \citep{kroegerLexicon2026}. This lexical model represents concepts, lemmas, and phonological forms and implements their associative retrieval. Its implementation also provides a symbolic readout of ordered syllables for selected lexical items, forming the interface to the articulatory pathway described here. The two contributions thus organize a common modeling program from concepts to movements using the same spiking neural modeling framework (NEF-SPA). In addition, an acoustic extension of DYNARTmo is described in a separate paper \citep{kroegerAcoustics2026}. This extension could use the articulatory trajectories generated here as input for speech synthesis.

The paper is organized as follows. The model description first introduces syllable programs and their lexical interface, followed by neural timing, gestural coordination, and articulatory realization. Separate sections describe the task performed with this architecture and the input corpus and simulation corpus used for the results. The discussion relates the implementation to earlier work and considers its extension toward integrated concept-to-acoustics simulation.

The primary aim of this paper is to provide a transparent description of the model, supported by illustrative simulations that can be explored in the \href{https://www.speechtrainer.eu/phono2gesture-pathway/}{visual simulation player}. A further aim is to examine whether prescribed gestures are recruited reliably across neural realizations. We report two paired simulation subcorpora of sixty runs each, using different gesture command amplitudes to drive the neural ramp integrators. This comparison identified settings that supported successful activation of all prescribed gestures in all sixty tested neural realizations. It complements the illustrative reference recordings across the input corpus and provides a basis for the broader evaluations outlined in Section~\ref{sec:evaluation}.

%% file: 02-representation.tex
\section{From lexical representations to syllable programs}
\label{sec:representation}
\subsection{Components and interfaces}
\begin{figure}[htbp]
\centering
\includegraphics[width=\linewidth]{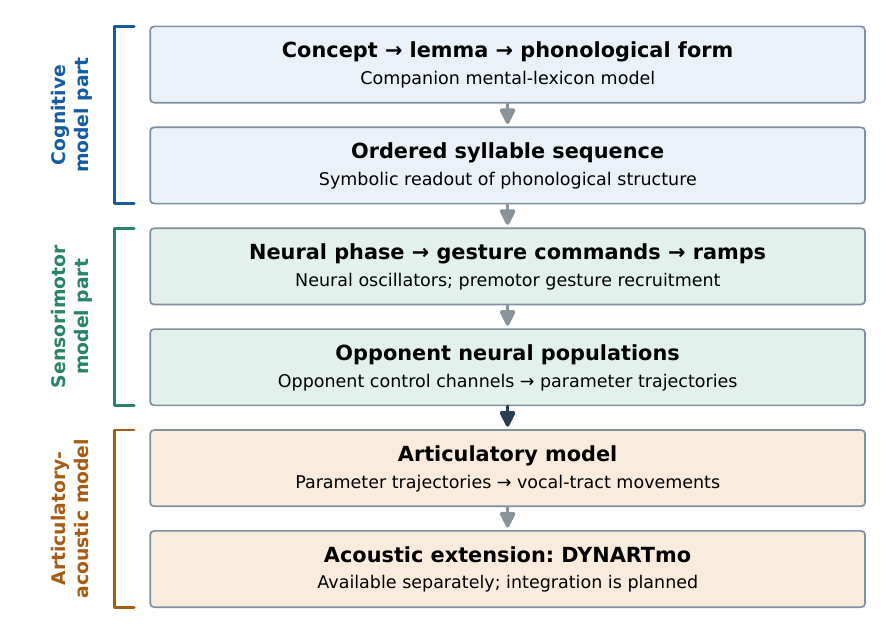}
\caption{Functional parts of the production pathway. Vertical labels and brackets group cognitive processing through ordered syllable readout, sensorimotor processing through opponent neural control, and the articulatory-acoustic model. The ordered syllable sequence forms the cognitive-to-sensorimotor interface; decoded parameter trajectories form the sensorimotor-to-articulatory interface. Gray arrows mark interfaces implemented through symbolic readout, external program selection, or numerical transfer, and the planned onward acoustic connection. The dark arrow from opponent outputs to geometry marks the direct implementation in the exporter. The diagram describes the functional components and their interfaces; details of the numerical execution are given in \ref{app:execution}. Schematic generated programmatically with AI-assisted code.}
\label{fig:architecture}
\end{figure}

Figure~\ref{fig:architecture} shows three functional components: a cognitive model part, a sensorimotor model part, and an articulatory-acoustic model. The cognitive model part retrieves a phonological form from a concept via a lemma. A symbolic readout then associates an accepted lexical item with an ordered list of syllables. This sequence forms the interface to the sensorimotor model part, where each syllable occurrence receives a neural oscillator and a set of gesture occurrences. The sensorimotor model part spans higher-level premotor gesture organization through the opponent neural populations that supply decoded articulatory control parameters. The articulatory model converts these parameter trajectories into vocal-tract positions and movements; its separately available acoustic extension provides the onward connection to sound.

Two levels of description are useful here. A syllable program specifies which syllables and gestures are to be realized, including gesture targets and their relative timing. The neural populations realize the temporal activity through which that knowledge is expressed. The following sections explain how a phase-dependent command becomes a smooth gesture activation and how several gestures jointly influence an articulator.

The current implementation connects these stages through explicit interfaces. The lexical readout accesses stored graph structure, and the corresponding sensorimotor syllable program is selected externally. Gesture activity supplies the inputs to the opponent neural populations, whose outputs are converted into vocal-tract movements by a deterministic geometric calculation. Section~\ref{sec:task} describes this functional pathway, \ref{app:execution} documents its numerical execution, and Section~\ref{sec:integration} discusses integration with lexical processing and acoustics.

\subsection{Ordered syllable readout}
The lexical resource stores phonological structures as relational graphs \citep{kroegerLexicon2026}. For the eight supported words, the readout follows the ordered links at the word node and extracts the syllables and their stress information. Syllable order is therefore available explicitly in the stored lexical structure. The readout uses these graph relations as the source of ordering information; the companion model's semantic pointers support lexical retrieval and similarity-based processing.

The readout applies a small set of documented boundary adjustments to obtain the implemented syllable programs. For the five words containing medial consonants commonly analyzed as ambisyllabic, we set aside their dual syllable affiliation at the gesture-generation interface. To realize each medial consonant once, we assign its gestures to the onset of the second syllable and omit the final consonant from the first stored syllable's output form. This convention is consistent with the onset assignment adopted by \citet[pp.~49--52, 61]{hall1992}. For example, the first stored form \code{kan} becomes \code{ka}, followed by \code{ne}, for \emph{Kanne}. These adjustments concern the readout of identified graph nodes and preserve the underlying lexical data. They specify the interface convention for the present examples; general syllabification beyond these examples is an extension discussed in Section~\ref{sec:integration}. The onset assignment remains compatible with temporal overlap between the medial consonant and the preceding vowel. Adjusting the initiation time of the following syllable's neural oscillator provides a way to vary this overlap and investigate German syllable cut (Section~\ref{sec:evaluation}).

\subsection{Syllable types and their occurrences}
A syllable program specifies an onset, a vocalic nucleus, and an optional coda: the consonantal material before the vowel, the vowel itself, and any consonantal material after it. Each consonantal position selects an oral constriction gesture, together with a glottal-opening gesture for a voiceless consonant or a velum-lowering gesture for a nasal consonant. The model's neutral setting corresponds to the articulatory configuration of a voiced, non-nasal schwa. In this simplified model, voiced sounds therefore require no separate voicing gesture, and non-nasal sounds require no separate velum-raising gesture. A nucleus selects one or two vowel gestures. The implemented structures include CV, CVC, CCV, CCVC, CVCC, and CCVCC, where C denotes a consonantal position and V a vocalic nucleus. Affricates represented by \code{ts} and \code{pf} are decomposed into two oral actions in these programs.

In the present NEF-SPA implementation, a syllable \emph{type} specifies a neural circuit configuration comprising a neural oscillator, phase-dependent gesture commands, and the associated gesture-activation populations. The neural oscillator design is shared across types; the selected gestures and their phase windows determine the syllable-specific coordination. The same circuit configuration can be instantiated for repeated occurrences of a type. In the reported simulations, each syllable \emph{occurrence} receives its own neural oscillator and gesture-activation populations. Thus, the first and fourth syllables in \code{bap--gum--dip--bap} use the same type but retain separate neural states. This preserves their individual timing until their effects are combined at the articulatory-parameter stage. Reusing the same neural populations across successive occurrences, and sharing coordination patterns across broader classes of syllables, are considered in Section~\ref{sec:syllable-reuse}.

A lexical item can also support alternative articulatory syllable programs. For \emph{Teller}, the lexical readout \code{tE--lEA} is realized either with one vowel gesture in the second syllable, yielding the short-ER program \code{tE--lA}, or with two overlapping vowel gestures, yielding the long-ER program \code{tE--lEA}. In both cases the second syllable has one neural oscillator. The codes identify the implemented realizations and are explained in \ref{app:codes}; the corresponding recordings are considered in Section~\ref{sec:variants}. The alternative realizations can also be explored in the \href{https://www.speechtrainer.eu/phono2gesture-pathway/}{visual simulation player}.

%% file: 03-neural-timing.tex
\section{Neural timing and gestural coordination}
\label{sec:timing}
\subsection{Representation and transformation in spiking populations}
\label{sec:neural-representation}
A neural population is a group of neurons whose combined activity represents a variable. In the sensorimotor model part, that variable can be the two-dimensional state of a neural syllable oscillator or a single value describing the strength of a gesture. These continuously changing variables are represented using the same NEF principles that support higher-dimensional semantic pointers in the SPA \citep{eliasmith2003,eliasmith2013}.

For a represented variable $\mathbf{x}$, the input current to neuron $i$ is
\begin{equation}
J_i=\alpha_i\mathbf{e}_i^{\mathsf T}\mathbf{x}+b_i.
\end{equation}
The preferred direction $\mathbf{e}_i$ determines which component of $\mathbf{x}$ excites the neuron; the gain $\alpha_i$ and bias $b_i$ determine its sensitivity and baseline input. Leaky integrate-and-fire (LIF) dynamics convert this current into a spike train $s_i(t)$. A readable estimate of the represented variable is recovered by weighting the filtered spike trains,
\begin{equation}
\widehat{\mathbf{x}}(t)=\sum_i\mathbf{d}_i\,(h*s_i)(t),
\end{equation}
where $\mathbf{d}_i$ is a decoding weight and $h*s_i$ denotes temporal filtering. A connection can similarly use decoding weights for a function of $\mathbf{x}$, thereby implementing a transformation between populations. Nengo constructs these connections from the specified functions and neural parameters \citep{bekolay2014}.

Filtering has two roles. A synaptic filter shapes the signal transmitted within the network and contributes to its dynamics. A probe filter smooths a recorded estimate for observation. Both make population activity accessible on a continuous time scale; only the former participates in the ongoing network computation. The recording settings and the interpretation of the resulting curves are introduced with the task in Section~\ref{sec:recording}.

\subsection{A neural oscillator as a temporal coordinate}
Each syllable occurrence has a population of 2,500 LIF neurons representing a two-dimensional state $\mathbf{x}_s=(x_s,y_s)^{\mathsf T}$. The two coordinates describe its position in a plane. As the state rotates around the origin, its angle provides a continuously advancing indication of progress through the syllable. For a desired frequency $f$, define
\begin{equation}
A=2\pi f\begin{pmatrix}0&-1\\1&0\end{pmatrix},
\qquad T=I+\tau A.
\end{equation}
Here, $A$ specifies rotation and $I$ is the identity matrix. The recurrent connection feeds the population output back to its input using the transform $T$ and a first-order low-pass synapse with $\tau=25$~ms. In the ideal decoded-state approximation, without an external input,
\begin{equation}
\tau\dot{\mathbf{x}}_s=-\mathbf{x}_s+Tz(\mathbf{x}_s).
\label{eq:oscillator}
\end{equation}
While the feedback function $z$ returns its input, Eq.~\eqref{eq:oscillator} reduces to $\dot{\mathbf{x}}_s=A\mathbf{x}_s$. The state rotates with nominal frequency $f=1.5$~Hz. The implementation suppresses recurrent feedback when $\operatorname{atan2}(y_s,x_s)>15\pi/16$, causing the activity to decay near the end of the cycle. Each initiation therefore produces an approximately one-cycle temporal process followed by a return toward inactivity.

The normalized phase is
\begin{equation}
p_s=\frac{\operatorname{atan2}(y_s,x_s)}{2\pi}+\frac12,
\qquad 0\leq p_s\leq1.
\label{eq:phase}
\end{equation}
The feedback cutoff corresponds to $p_s>31/32$, or $348.75^\circ$ on the displayed phase scale. A negative input along the first coordinate starts the state near the beginning of this scale. The first occurrence is initiated by the task input; later occurrences receive their initiation signal from the preceding neural oscillator.

This construction explains the role of neural oscillators in the model. A gesture beginning at phase $a$ and ending at phase $b$ has an ideal command duration
\begin{equation}
\Delta t=\frac{b-a}{f}.
\label{eq:duration}
\end{equation}
For example, a phase interval of 0.15 cycles ($54^\circ$) corresponds to 100~ms at 1.5~Hz. Relative intervals thus become durations through the recurrent dynamics. The syllable program specifies each gesture relative to the shared phase, and the evolving neural state determines when that phase is reached. Frequency, phase intervals, and synaptic constants define this temporal mechanism; finite-population approximation, initiation transients, and filtering contribute to the recorded trajectories.

\subsection{Phase-dependent gesture commands and syllable initiation}
A gesture command indicates the part of the syllable cycle during which a particular action is recruited. Gesture occurrence $g$, assigned to syllable occurrence $s$, has a prescribed start phase $p_g^{\mathrm{on}}$, end phase $p_g^{\mathrm{off}}$, and gesture command amplitude $c_g$. Its desired command is
\begin{equation}
u_g(\mathbf{x}_s)=
\begin{cases}
c_g,&\norm{\mathbf{x}_s}\geq0.3\ \text{and}\ p_g^{\mathrm{on}}\leq p_s<p_g^{\mathrm{off}},\\
0,&\text{otherwise}.
\end{cases}
\label{eq:window}
\end{equation}
The amplitude condition restricts command generation to an active neural oscillator whose phase provides a useful temporal reference. In networks with preselected syllables, a decoded connection approximates Eq.~\eqref{eq:window} and drives a 200-neuron pulse population. This population represents the short-lived command that subsequently recruits the gesture activation population. Different gestures read different intervals of the same phase. Their relative organization is therefore specified within the syllable program while their execution follows the neural state.

For consecutive syllables, the interval $0.45$ ($162^\circ$) $\leq p_s<0.50$ ($180^\circ$) generates a next-slot command of amplitude 1.5. In networks with preselected syllables, a separate 200-neuron population forwards this command to neural oscillator $s+1$. The approximately one-cycle activity of successive neural oscillators therefore overlaps substantially in time, with each neural oscillator carrying the temporal state of its own syllable occurrence.

This early initiation allows the next neural oscillator to build up activity and establish its phase trajectory before recruiting its gestures. The model reserves the initial part of the cycle for this startup: the first oral onset command begins at $p_{s+1}=0.15$ ($54^\circ$), corresponding to a nominal phase advance of 100~ms at 1.5~Hz. The observed delay also depends on initiation transients and synaptic filtering. Together, early initiation and delayed gesture recruitment coordinate the following onset with the later gestures and offset of the preceding syllable. We interpret this startup interval as a syllable preparation time within the model. Its neurobiological interpretation is the hypothesis that preparatory neural activity for a syllable precedes recruitment of its articulatory gestures.

The syllable-specific information required here belongs to our proposed mental syllabary (gesture identities and phase values defining the start and end of each command window; Section~\ref{sec:syllable-reuse}). The NEF implementation uses these specifications to transform neural oscillator phase into commands for the neural ramp integrators. SPA mechanisms organize the selection of syllable identities and sequence positions through semantic pointers. This selection and the underlying timing circuits are implemented within the same NEF-based spiking neural framework.

The SPA-controlled configuration, adapted from the earlier model \citep{kroeger2023}, separates two control functions. First, the sequence controller initiates the neural oscillator assigned to the next position; the reported sequence uses four neural oscillators of the same design. Second, as each neural oscillator becomes active, a controller for that position retrieves the associated syllable identity from the stored syllable--position representation. This retrieval enables the corresponding gesture pathways. Inhibitory populations normally suppress the relay populations receiving syllable-specific phase commands. Retrieval of the selected syllable reduces this inhibition, a process termed \emph{disinhibition}, allowing the active neural oscillator's phase commands to reach the appropriate neural ramp integrators through the released pathways. The gesture identities and phase windows are already encoded in these connections. Selecting them determines which gestures the neural oscillator recruits and at which phases; a separate initiation pathway starts the neural oscillator itself.

\subsection{Relative coordination of the gestures}
The current phasing rules define oral, vocalic, glottal, and velopharyngeal gesture duration intervals on a common syllable phase scale. Their specification follows a hierarchy. For the basic syllable pattern, the vocalic gesture is first placed around the middle of the neural oscillator cycle from $0^\circ$ to $360^\circ$. Oral constriction gestures are then positioned relative to it for the onset and, where present, the coda. Additional oral constriction gestures are placed for a second consonant in an onset or coda cluster. Finally, the glottal-opening gesture of a voiceless consonant or the velum-lowering gesture of a nasal is phased relative to that consonant's oral constriction gesture.

These phasing rules form an initial, heuristic specification. Their organization draws on the rule-based construction of gesture scores in DYNARTmo described by \citet[Section~3]{kroegerGestures2025}. The numerical phase values used here are provisional choices informed by the author's prior experience of adjusting gesture coordination in articulatory synthesis to obtain perceptually plausible speech. Quantitative fitting to natural articulatory data remains a next step.

\begin{figure}[htbp]
\centering
\includegraphics[width=\linewidth]{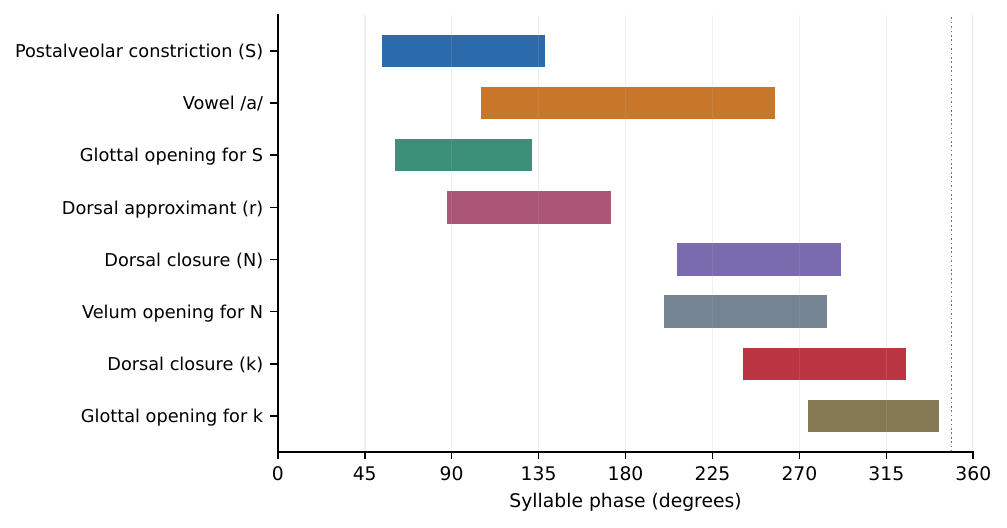}
\caption{Gesture activation windows for \emph{Schrank} (SAMPA transcription: \code{SraNk}). The oral, glottal, vocalic, and velopharyngeal actions occupy overlapping phase intervals. The dorsal actions for \code{N} and \code{k} are separate occurrences even though they share a constriction target. Bar height is for display only; amplitudes differ by gesture class. The dotted line marks the neural oscillator feedback cutoff. Schematic generated programmatically with AI-assisted code.}
\label{fig:phase}
\end{figure}

Table~\ref{tab:phase} gives the principal time windows. A single onset consonant starts at $54^\circ$. For a two-consonant onset, the second oral action starts approximately $33.80^\circ$ after the first. The vowel starts at approximately $105.37^\circ$, while coda gestures start later in the same cycle. For example, the onset command can remain active after the vowel command begins, so their temporal overlap is already specified at the program level. Figure~\ref{fig:phase} illustrates these prescribed relationships for \emph{Schrank}; the resulting neural activation trajectories are examined after the task and input corpus have been introduced. The \href{https://www.speechtrainer.eu/phono2gesture-pathway/}{visual simulation player} provides a complementary view of gesture timing and articulatory movement.

\begin{table}[htbp]
\centering\small
\caption{Prescribed gesture activation windows in normalized syllable phase, with degrees in parentheses. Both representations are rounded independently from the full-precision source values. The intervals define the phase commands from which neural activation and movement subsequently develop.}
\label{tab:phase}
\begin{tabular}{lrr}
\toprule
Gesture position & Start & End\\
\midrule
Single onset consonant & 0.1500 ($54.00^\circ$) & 0.4317 ($155.41^\circ$)\\
Two-consonant onset: first & 0.1500 ($54.00^\circ$) & 0.3848 ($138.51^\circ$)\\
Two-consonant onset: second & 0.2439 ($87.80^\circ$) & 0.4787 ($172.32^\circ$)\\
Stressed single vowel & 0.2927 ($105.37^\circ$) & 0.7152 ($257.49^\circ$)\\
Unstressed single vowel & 0.2927 ($105.37^\circ$) & 0.6683 ($240.59^\circ$)\\
Two-vowel nucleus: first & 0.2927 ($105.37^\circ$) & 0.6213 ($223.68^\circ$)\\
Two-vowel nucleus: second & 0.5744 ($206.78^\circ$) & 0.8561 ($308.20^\circ$)\\
Single coda consonant & 0.5744 ($206.78^\circ$) & 0.8561 ($308.20^\circ$)\\
Two-consonant coda: first & 0.5744 ($206.78^\circ$) & 0.8091 ($291.29^\circ$)\\
Two-consonant coda: second & 0.6683 ($240.59^\circ$) & 0.9030 ($325.10^\circ$)\\
Initiation of next syllable & 0.4500 ($162.00^\circ$) & 0.5000 ($180.00^\circ$)\\
\bottomrule
\end{tabular}
\end{table}

Glottal opening is coordinated relative to its oral gesture. For a voiceless plosive, glottal command onset is delayed by 0.09390 cycles and offset by 0.04695 cycles relative to oral onset and offset. For a voiceless fricative, the glottal interval begins 0.01878 cycles later and ends 0.01878 cycles earlier than the oral interval. At the nominal frequency, these phase differences correspond respectively to approximately 62.6, 31.3, and 12.5~ms. These values specify the relative commands for oral and glottal coordination. Their relationship to acoustic landmarks can be evaluated after acoustic synthesis has been connected to the model (Section~\ref{sec:integration}).

Velum opening is associated with nasal gestures. Its command normally begins 0.01878 cycles before the nasal oral command and ends the same amount later. If an obstruent immediately precedes or follows within the same consonant cluster, the corresponding boundary is moved inward instead. This phasing rule adjusts nasal overlap with adjacent oral actions while retaining their common phase reference. The earlier ending of an unstressed single-vowel command provides one limited stress-related timing contrast; broader prosodic control is a further extension (Section~\ref{sec:evaluation}).

\subsection{From phase commands to activation ramps}
\label{sec:gesture-ramps}
A brief command drives a neural ramp integrator that generates a gesture's activation function, including its rise and fall over time. This smooth response is called an \emph{activation ramp}. Each gesture occurrence has a separate recurrent population of 1,000 LIF neurons representing its activation $r_g$. A pulse drives this population with the ramp rate factor $v_g$ as its input gain, and recurrent feedback has gain $1-v_g$ with the function $\ell(r)=\min(0.9,r)$. Both current configurations use $v_g=0.05$ for vowels and velum opening, and $v_g=0.20$ for the other gestures. With the default 5-ms connection synapses, an ideal approximation is
\begin{equation}
\tau_r\dot r_g=-r_g+(1-v_g)\ell(r_g)+v_gu_g,
\qquad \tau_r=5\ \mathrm{ms}.
\label{eq:ramp}
\end{equation}
Below the feedback ceiling, this becomes $\tau_r\dot r_g=v_g(u_g-r_g)$. The corresponding effective time constants are 100~ms and 25~ms. This circuit acts as a leaky, saturating tracker of a command, rather than an unbounded mathematical integrator; the term ramp integrator describes its functional role in shaping gesture recruitment. These values explain the slower rise and decay of vocalic and velopharyngeal activity relative to oral-constriction and glottal activity. The neural realization additionally reflects approximation in the pulse and ramp populations, while the observation filter shapes the recorded estimate (Section~\ref{sec:recording}).

The gesture command specifies recruitment within a phase region; the activation describes the resulting population response. Articulatory movement follows when that response is combined with the gesture target, as described in Section~\ref{sec:articulation}. A command may end while activation is still decaying, allowing effects from one gesture to persist into a subsequent gesture or syllable.

\emph{Gesture command amplitude} $c_g$ and \emph{ramp rate factor} $v_g$ are distinct parameters (Table~\ref{tab:ramp-parameters}). The former specifies the level of the command during its phase window; the latter weights feedforward drive and, through $1-v_g$, recurrent feedback. In the unsaturated idealization, a constant command $c_g$ starting from zero gives
\begin{equation}
r_g(t)=c_g\left(1-e^{-v_gt/\tau_r}\right).
\label{eq:ramp-step}
\end{equation}
Thus $v_g$ controls the response time, while both $c_g$ and $v_g$ affect the initial slope $v_gc_g/\tau_r$. A finite command window may end before its level is approached. The actual pulse, routing, and ramp populations add decoding and transmission effects to this approximation. Neither $v_g$ nor $c_g$ is a physical movement velocity, and decoded gesture trajectories are not individually rescaled to a peak of one before target weighting.

\begin{table}[!htbp]
\centering\small
\caption{Gesture command amplitude and ramp rate factor. Amplitudes $c_g$ use the dimensionless 0--1 reference scale defined in Section~\ref{sec:amplitude-reference}. Two sets of gesture command amplitude values are used to generate the paired simulation subcorpora to investigate parameter-sensitive activation failures (Section~\ref{sec:paired-results}). For syllable-coda glottal opening, the original amplitude is 0.60 in \code{bap--gum--dip--bap} and 1.00 in \emph{Schrank}. The ramp rate factor is identical in both subcorpora.}
\label{tab:ramp-parameters}
\begin{tabular}{lrrr}
\toprule
Gesture class & Original $c_g$ & Revised $c_g$ & $v_g$\\
\midrule
Oral consonantal & 1.00 & 1.00 & 0.20\\
Vocalic & 0.95 & 0.95 & 0.05\\
Velum opening & 0.70 & 0.90 & 0.05\\
Glottal opening (syllable onset) & 1.00 & 0.70 & 0.20\\
Glottal opening (syllable coda) & 0.60 / 1.00 & 0.70 & 0.20\\
\bottomrule
\end{tabular}
\end{table}

The value 0.9 in $\ell(r)$ limits the decoded recurrent feedback function; it is neither the gesture command amplitude nor a hard bound on the observed population readout. Raw decoded activity can also transiently become negative. Such excursions are retained in the plots, while the articulatory target interface uses bounded nonnegative gesture weights (Section~\ref{sec:articulation}).

\subsection{Network configuration and random seeds}
\label{sec:configuration}
The size of a network with preselected syllables follows from its syllable and gesture occurrences. For $S$ syllable occurrences and $G$ gesture occurrences, the timing network contains
\begin{equation}
N_{\mathrm{program}}=2500S+200(S-1)+1200G
\label{eq:count}
\end{equation}
neurons: 2,500 per neural oscillator, 200 for each next-syllable command, and 200 pulse plus 1,000 ramp neurons per gesture. The opponent populations introduced in Section~\ref{sec:opponents} form the output stage of the sensorimotor model part. The adapted SPA-controlled sequence additionally includes its controller and routing circuitry, bringing its timing and selection network to 309,820 neurons.

A random seed is an integer used to initialize a pseudorandom number generator, fixing the sampled choices used to construct a neural realization.\footnote{See the \code{seed} parameter of \code{nengo.Network} in the Nengo documentation: \url{https://www.nengo.ai/nengo/v4.0.0/frontend-api.html\#nengo.Network} (accessed 7 November 2026).} In both configurations, these choices concern the NEF-based neural populations, including their tuning parameters: preferred directions and the points at which neurons begin responding to their represented input. With the same model construction and software configuration, keeping the seed fixes these choices; changing it gives another neural realization of the specified computation. The illustrative networks with preselected syllables use network seed 11; the paired simulation subcorpora include runs with different neural seeds as specified in Section~\ref{sec:paired-protocol}. Their phase windows and articulatory targets are explicitly assigned production knowledge and remain fixed when this neural seed is changed.

At the SPA level, the controller additionally requires a specification of the semantic-pointer vocabulary vectors representing syllable identities and sequence positions. A seed for the neural network and a specification of those vocabulary vectors concern different parts of the model. The reproducibility of the adapted controller is discussed in Section~\ref{sec:earlier}. \ref{app:parameters} collects the neural parameters; the task timing and recording settings are specified separately in Section~\ref{sec:task}.

\subsection{Reference amplitudes in neural ensembles and SPA buffers}
\label{sec:amplitude-reference}
In this model, a value of one provides a common reference for represented activation strength in neural ensembles and SPA buffers. For nonnegative activation signals, values near zero indicate weak or absent represented activation, and values near one indicate reference-strength activation. These population-level values are distinct from individual neuronal firing rates; a represented zero can coexist with ongoing spiking. NEF-SPA also supports signed variables, such as the neural oscillator coordinates, and model-specific representational scales \citep{bekolay2014}.

At the SPA level, the elementary vocabulary vectors are generated with Euclidean length one. This normalization concerns the length of the entire vector; its individual components can be positive or negative \citep{eliasmith2013}. For a unit vector $\mathbf{p}$, a buffer state $a\mathbf{p}$ with $a\geq0$ has length $a$ and represents that pointer at the corresponding strength. Adding pointers can increase vector length. Composite vectors can be explicitly renormalized when a unit-length representation is required; the buffer state need not have unit length throughout processing.\footnote{See the NengoSPA documentation for unit-length vocabulary generation and explicit normalization of composed pointers: \url{https://www.nengo.ai/nengo-spa/v1.2.0/modules/nengo_spa.vocabulary.html} (accessed 8 November 2026).}

Commands use the same unit reference. The gesture command amplitudes $c_g$ used here lie between 0 and 1: 0 denotes no command, and 1 the full reference command. Thus, 0.70 means 70\% of that reference level. Stronger commands remain possible: the external input initiating the first neural oscillator has unit amplitude, whereas the phase-dependent next-syllable command has amplitude 1.5 (Sections~\ref{sec:timing} and~\ref{sec:task}). One is therefore a reference value rather than a universal ceiling. Neural dynamics and decoding can also produce activation values outside the nominal interval. The resulting gesture trajectories retain their decoded amplitudes rather than being individually rescaled to peak at one (Section~\ref{sec:gesture-ramps}); explicit bounds at the articulatory interface are described in Section~\ref{sec:articulation}.
\FloatBarrier

%% file: 04-articulation.tex
\section{From gesture activity to articulatory movement}
\label{sec:articulation}
Gesture occurrences retain separate activation trajectories as they pass from timing to articulatory control. This section describes how their contributions provide inputs to one pair of opponent neural populations per articulatory parameter, and how the decoded population outputs then drive the geometric articulatory model. When concurrently active gestures specify the same parameter, their contributions meet at the input to its opponent population pair.

\subsection{From gesture targets to opponent population inputs}
Gesture activation specifies how strongly an action contributes at a given time. Articulatory targets specify what that action contributes. The implementation associates each gesture occurrence with a subset of continuous parameters, including vowel height and frontness, lip rounding and closure, apical and dorsal constriction, velum position, and glottal aperture. Apical constriction concerns the tongue tip, dorsal constriction the tongue body, and velum movement the opening toward the nasal tract. A ninth parameter, apical frontness, distinguishes alveolar and postalveolar tongue-tip configurations in the networks with preselected syllables. Targets are drawn from the articulatory trainer's gesture definitions and are retained in each recording's metadata.

Concurrent gestures contribute to the parameters specified by their respective targets. For example, in the present target definitions, both a vowel gesture and an apical closure gesture specify vowel height and frontness. For each shared parameter, their decoded activations weight the respective target values. Positive and negative contributions are collected separately to form the inputs to its opponent neural populations (Section~\ref{sec:opponents}). The numerical implementation of this input mapping is documented in \ref{app:execution}.

Let $\mathcal G_p$ be the set of gestures that assign a target to parameter $p$, let $a_g=\operatorname{clip}(\widehat r_g,0,1)$ be the clipped decoded activation, and let $q_{gp}=\operatorname{clip}(q^{\mathrm{stored}}_{gp},-1,1)$ be the target used in the mixture. Define
\begin{equation}
W_p=\sum_{g\in\mathcal G_p}a_g,\qquad
w_{0p}=\max(0,1-W_p),\qquad D_p=\max(1,W_p).
\end{equation}
For a neutral value $q_{0p}$, positive and negative commands are
\begin{align}
d_p^+&=\frac{\displaystyle\sum_{g\in\mathcal G_p}a_g[q_{gp}]_+
                   +w_{0p}[q_{0p}]_+}{D_p},\\
d_p^-&=\frac{\displaystyle\sum_{g\in\mathcal G_p}a_g[-q_{gp}]_+
                   +w_{0p}[-q_{0p}]_+}{D_p},
\label{eq:opponent}
\end{align}
where $[x]_+=\max(x,0)$. Here, $W_p$ is the total activation weight for parameter $p$, $w_{0p}$ supplies any remaining weight from the neutral configuration, and $D_p$ keeps the combined command bounded. For a single positive target of one, activation 0.6 and a neutral value of zero give a positive command of 0.6. When total activation exceeds one, normalization produces a weighted average of the active targets. The mixture is parameter-specific: a vowel-height command combines the gestures that specify vowel height, while a glottal-opening command combines those that specify glottal aperture.

Negative decoded gesture samples are mapped to zero before calculating $W_p$ and the target contributions. They therefore generate no negative gesture weight and no active contribution from that gesture. This does not imply that the complete articulatory command is zero: neutral contributions and other gestures remain. It also does not remove negative internal states from the recurrent ramp dynamics. A negative gestural target, such as the velum-lowering direction, is different: it is weighted by a positive activation and legitimately contributes to $d_p^-$.

Splitting target signs before summation retains simultaneous positive and negative drive. For example, equally weighted targets of $+1$ and $-1$ can produce substantial activity in both channels even when their net command is zero. The two channels therefore preserve information about concurrent contributions as well as their net effect. They represent abstract opposing influences on a model parameter; their biological interpretation is considered in Section~\ref{sec:evaluation}.

\subsection{Opponent neural populations}
\label{sec:opponents}
The opponent neural populations form the final neural stage of the sensorimotor model part. An opponent pair contains two populations for one parameter: one represents drive in the positive direction and the other drive in the negative direction. Both population variables are nonnegative. This organization allows opposing contributions to remain visible before they are combined into a signed control value.

Each channel contains 300 LIF neurons and receives the corresponding command $d_p^+$ or $d_p^-$. Positive encoders make the neurons respond to increasing positive input. Their intercepts, sampled uniformly between zero and one, distribute the input levels at which individual neurons begin responding. Input connections transmit the commands directly, with no added synaptic filter. The networks with preselected syllables use nine parameters and therefore 18 populations, or 5,400 neurons. The adapted SPA-controlled sequence uses eight parameters and 4,800 neurons. Network seed 23 fixes the sampled neural realization of this stage.

After decoding and observation filtering, the channel readouts are bounded to $[0,1]$ to obtain $m_p^+$ and $m_p^-$. The raw decoded values are also retained for inspection. Their difference gives the signed control value
\begin{equation}
m_p(t)=m_p^+(t)-m_p^-(t).
\end{equation}
The parameter-wise commands defined above supply the inputs to these populations, whose decoded outputs then drive the geometric stage. Section~\ref{sec:task} describes the production procedure; \ref{app:execution} gives its computational implementation.

\subsection{Geometric realization and visualization}
The articulatory model begins at the interface from decoded neural control values to geometric parameters. It determines the position and shape of the visible articulators from these parameter trajectories. An adapter translates the normalized outputs of the sensorimotor model part into its parameters:
\begin{equation}
Q_p=o_p+g_p^+m_p^+-g_p^-m_p^-.
\label{eq:geometry}
\end{equation}
The offset $o_p$ sets the resting configuration, while the gains $g_p^+$ and $g_p^-$ set the effect of each control direction on geometry. Most gains are one. The adapter uses positive gains of 1.1 for apical constriction and 1.2 for dorsal constriction and glottal aperture, a negative gain of 1.3 for velum movement, and a resting offset of $+0.1$ for the velum. These explicit adjustments belong to the articulatory interface. In particular, stored constriction targets of 1.07 are clipped to one in the opponent mixture; the geometric gain is applied afterwards.

The resulting parameters control a Python port of DYNARTmo's contour interpolation \citep{kroegerGestures2025}. Vowel-related reference configurations are combined with lip, tongue-tip, tongue-dorsum, jaw, and velum adjustments.

A \href{https://www.speechtrainer.eu/phono2gesture-pathway/}{visual simulation player} displays these movements alongside the recorded neural activity. Links to the player, source code, and explanatory documentation are provided in Section~\ref{sec:resources}. In the player, the midsagittal view shows a section through the middle of the vocal tract; a separate superior view displays the glottis from above. This is a kinematic visualization of articulatory movement: each sample produces a configuration, and playback displays the successive configurations on their original time grid. Playback retains the recorded timing of the neural cascade.

The adapter exports the contours together with parameter trajectories and the original timestamps. A separate parameter file expresses the same commands in trainer units by multiplying normalized values by 1,000. Categorical information, such as full, fricative, or lateral constriction \citep{kroegerGestures2025}, remains attached to individual gestures where needed. These labels preserve information needed for the acoustic integration discussed in Section~\ref{sec:integration}, where the consequences of different constriction modes must also be realized.

The visible movement is thus the consequence of an explicit computational chain: phase-dependent neural commands generate gesture activity; the activity weights articulatory targets; opponent neural outputs determine control values; and those values determine contours. A displayed closure can be traced back through this chain to its syllable and gesture occurrence. This stage provides a kinematic realization, expressing how the geometry changes over time. Tissue mechanics and the aerodynamic and acoustic consequences of these changes are considered separately in the discussion.

%% file: task-realization.tex
\section{Task realization using the neural architecture}
\label{sec:task}
The task is to realize individual syllables and words represented by sequences of articulatory syllable programs as temporally organized patterns of gesture activity and articulatory movement. A trial starts with a selected word, syllable, or syllable sequence and ends with recorded control trajectories and synchronized vocal-tract movements. This procedure exercises the production stages described above.

\subsection{Preparation and initiation of a word or syllable sequence}
\begin{figure}[!htbp]
\centering
\includegraphics[width=\linewidth]{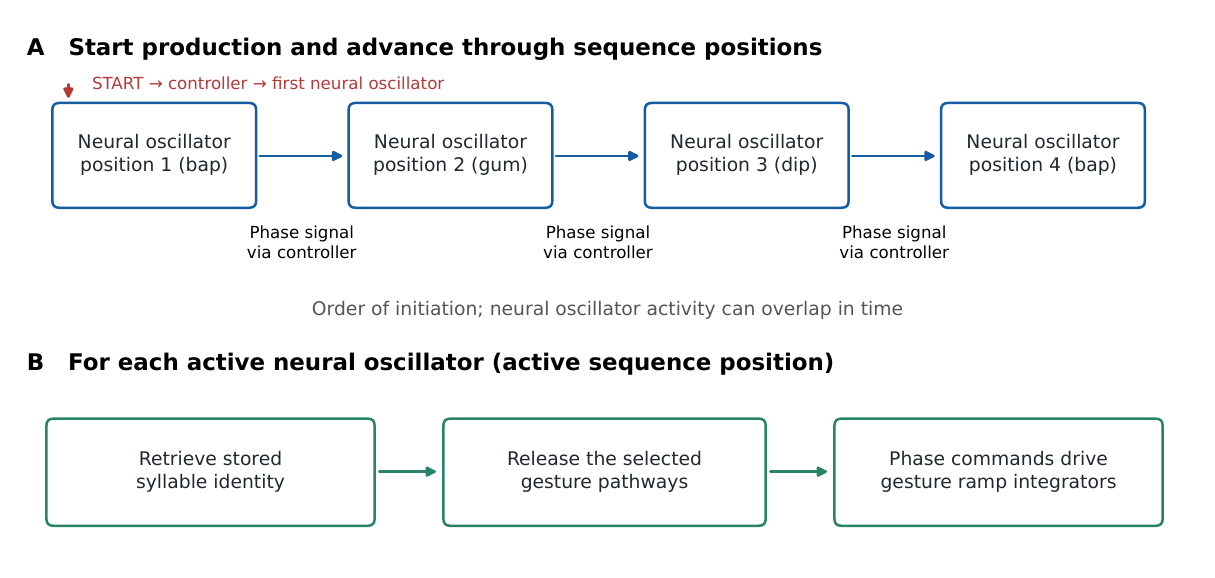}
\caption{Sequential initiation and gesture recruitment in the SPA-controlled four-syllable example. (A) The controller initiates the first neural oscillator and subsequently advances through the positions in response to next-syllable phase signals. The arrows indicate initiation order; neural oscillator activity can overlap in time. (B) For each active neural oscillator, retrieval of the syllable identity stored for its sequence position releases the appropriate gesture pathways, through which that oscillator's phase commands drive the gesture ramps. The diagram shows the functional sequence without assigning a duration to utterance preparation. The startup of each neural oscillator provides its syllable preparation time; overlapping neural oscillator activity is shown in Figure~\ref{fig:repeat}. Schematic generated programmatically with AI-assisted code.}
\label{fig:sequence-overview}
\end{figure}

This subsection describes how production of a selected word or syllable sequence is prepared and initiated. Figure~\ref{fig:sequence-overview} illustrates this procedure for a SPA-controlled four-syllable example representing a nonsense utterance (\code{bap--gum--dip--bap}). Preparation makes the syllable identities and their order available for production. The gesture specifications described in Sections~\ref{sec:representation} and~\ref{sec:timing} provide the associated production knowledge. Individual gestures are subsequently recruited during production, according to the evolving phase of the neural oscillator for the corresponding syllable.

First, the task supplies four syllable identities together with their sequence positions. For \code{bap--gum--dip--bap}, the pairs are \code{bap}/position~1, \code{gum}/position~2, \code{dip}/position~3, and \code{bap}/position~4 (Figure~\ref{fig:sequence-overview}). SPA vectors encode these syllable--position associations, making the upcoming syllables available for retrieval and distinguishing the two occurrences of \code{bap}. The positions specify sequence order; phase-dependent commands determine the subsequent neural oscillator initiation times during production. The loading schedule used in the reported simulations is a prescribed input protocol (\ref{app:task-settings}), rather than a model-derived duration of utterance preparation.

With the upcoming syllables available, an external start command switches the sequence controller to production and initiates the neural oscillator for position~1. This begins the syllable preparation time described in Section~\ref{sec:timing}: the startup of the individual neural oscillator before gesture recruitment. For each active neural oscillator, the model retrieves the syllable identity stored for its sequence position and reduces inhibition on the corresponding gesture pathways. The oscillator's evolving phase can then recruit that syllable's gesture ramp integrators through those pathways.

Production proceeds sequentially. When an active neural oscillator reaches the next-syllable phase window, its signal causes the controller to initiate the neural oscillator for the following position. The same retrieval and pathway-selection process then operates for the newly active neural oscillator. Successive neural oscillators overlap in time, allowing preparation of the next syllable while gestures of the preceding syllable are still active (Figure~\ref{fig:repeat}). The external start command thus launches the production sequence; the phase-dependent signals organize its subsequent progression.

For longer utterances, our working hypothesis is a preparation window containing at most three or four upcoming syllables. This window would be updated as production advances, making further syllable identities and their order available while phase-dependent commands continue to initiate successive neural oscillators. The reported four-syllable example implements sequential release from a supplied sequence; continual updating of the preparation window remains to be implemented and evaluated.

\begin{figure}[!htbp]
\centering
\includegraphics[width=0.52\linewidth]{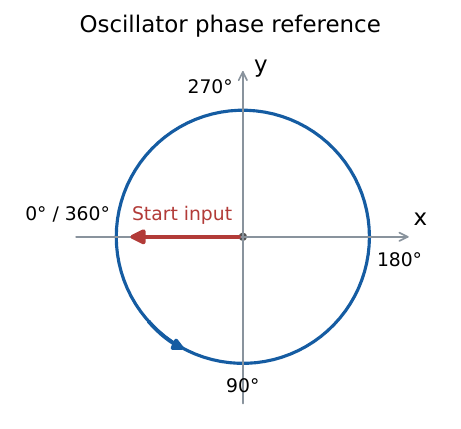}
\caption{Phase reference for a neural syllable oscillator. The schematic shows the oscillator's two-dimensional state space, using the same convention as the \href{https://www.speechtrainer.eu/phono2gesture-pathway/}{visual simulation player}: $0^\circ/360^\circ$ lies on the negative $x$ axis, and phase increases counterclockwise. The red arrow indicates the direction of the initiating input to the first coordinate $x$. The circle is a phase reference, not a recorded neural trajectory. Schematic generated programmatically with AI-assisted code.}
\label{fig:oscillator-start}
\end{figure}

\begingroup\clubpenalty=10000
A 50-ms input pulse, applied from 0.15 to 0.20~s, initiates the neural oscillator representing the first syllable to be produced in networks with preselected syllables. It has unit amplitude and acts in the negative direction of the neural oscillator's first coordinate (Figure~\ref{fig:oscillator-start}). Subsequent syllables are initiated by the phase-dependent next-syllable commands described in Section~\ref{sec:timing}. After the initial pulse, the network's own dynamics carry the program forward.\par\endgroup

The timing and gesture network is simulated in numerical integration steps of 1~ms, with a common recording window covering the interval from the first preparation action through production of the selected utterance to the final decay of activity (\ref{app:task-settings}). Gesture activation determines the positive and negative inputs to the opponent neural populations through the mapping described in Section~\ref{sec:articulation}. Their decoded outputs provide the articulatory control trajectories from which the geometric adapter generates synchronized vocal-tract movements. \ref{app:execution} documents the numerical execution of these stages in the reported simulations and the planned online implementation.

\subsection{Recording neural activity and movement}
\label{sec:recording}
The neural signals are recorded using 10-ms low-pass probe filters. This modest temporal averaging makes the represented population variables easier to follow by reducing rapid fluctuations in their decoded readout. It leaves the ongoing network computation unchanged. The filter on the gesture readout also forms part of the input pathway to the opponent populations (\ref{app:execution}). Precise timing measurements consequently refer to these recorded signals and settings.

The exports used for the illustrative figures and for playback in the \href{https://www.speechtrainer.eu/phono2gesture-pathway/}{visual simulation player} (Section~\ref{sec:resources}) retain one sample every 5~ms: 500 samples from 0.001 to 2.496~s. Neural oscillator coordinates, syllable-initiation signals, separate gesture activations, opponent inputs and outputs, control parameters, and contour frames share this recording grid. The illustrative figures use these stored samples directly; the paired simulation subcorpora additionally retain the upstream 1-ms traces used to evaluate activation reliability (Section~\ref{sec:paired-protocol}). The neural activation trajectories, as well as the resulting geometry of the vocal-tract configurations, can therefore be inspected at corresponding timestamps throughout the recording.

\subsection{Descriptive measurements}
\label{sec:measurements}
The analysis describes sequencing, gesture activity, and overlap within each recorded realization. Syllable initiation is marked by the first retained sample at which its recorded initiation signal exceeds 0.4. In networks with preselected syllables these signals are the filtered first input pulse and subsequent next-syllable pulses. In the SPA-controlled sequence they are decoded similarities to the syllable-position initiation symbols. The threshold provides a descriptive marker within each configuration; it marks initiation activity, from which gesture activation and movement subsequently develop.

Gesture overlap is measured by counting samples at which both selected gesture activations exceed 0.2 and multiplying by the 5-ms sample interval. This convention measures the duration occupied by the retained samples; the difference between the first and last sample times is one interval shorter. For each recording, the analysis also reports the smallest peak among its gesture traces, showing whether all prescribed gestures became active. Settling is described by the largest absolute mean of any gesture over the final 20 retained samples, a 100-ms observation window. These measures summarize the behavior of the implemented programs. Evaluation against human movement and acoustic data is discussed in Section~\ref{sec:evaluation}.

%% file: simulation-corpus.tex
\section{Input corpus and simulation corpus}
\label{sec:corpus}
\subsection{Input corpus and simulation subcorpora}
\label{sec:corpus-organization}
We distinguish the \emph{input corpus}, the inventory of selected words, nonwords, syllables, and syllable sequences, from the \emph{simulation corpus}, the collection of recorded model realizations analyzed in this paper. A \emph{simulation run} is one execution of the model for a given input sequence with specified model parameters and a neural seed. The input corpus comprises 30 input selections: eight lexical words with an additional realization variant of one word, 20 isolated syllables or syllabic parts of words, and one four-syllable nonword sequence in the SPA-controlled configuration. Some selections reuse syllables or sequences appearing elsewhere in the input corpus.

The illustrative reference runs cover the entire input corpus, with one saved recording per selection, yielding 30 recordings. Recorded examples can be explored in the \href{https://www.speechtrainer.eu/phono2gesture-pathway/}{visual simulation player} on the SpeechTrainer website (Section~\ref{sec:resources}), which displays neural activity alongside synchronized vocal-tract movements. Two additional paired \emph{simulation subcorpora} each contain recordings from 60 runs over the same three items of the input corpus: two words and the four-syllable sequence. These subsets of the simulation corpus repeat the selected inputs across neural seeds under the original and revised gesture command amplitudes, respectively. Their paired comparison assesses which of these parameter settings support reliable gesture activation (Section~\ref{sec:paired-protocol}).

The existing illustrations are retained as a fixed reference, rather than replaced by the current \href{https://www.speechtrainer.eu/phono2gesture-pathway/}{web-player recordings}. They demonstrate individual neural realizations and need not match a later playback selection. These illustrative reference recordings use the phasing specification \code{2026-11-05-onset54}, dated 5 November 2026, with the model configuration and procedure described above. The networks with preselected syllables use Nengo 4.1.0 and network seed 11; the opponent stage uses seed 23. The figures and descriptive measurements in the following section are calculated from these saved recordings. The retained data and their provenance provide a fixed reference within the simulation corpus for inspecting the model and for extending the evaluation to further programs and neural realizations.

\subsection{Paired checks across neural realizations}
\label{sec:paired-protocol}
In addition to the 30 illustrative reference runs, we evaluated two paired simulation subcorpora of 60 runs each. Each subcorpus comprised 20 realizations of \emph{Kanne} (neural seeds 101--120), 20 of \emph{Schrank} (201--220), and 20 of \code{bap--gum--dip--bap} (301--320). The runs in both subcorpora used the same seeds, input sequences, phase windows, population sizes, and articulatory targets. For the SPA-controlled sequence, the semantic-pointer vocabulary was held fixed using seed 90261106; its vectors were retained and verified to be identical across paired runs. The opponent stage retained seed 23.

The original gesture command amplitude was 0.70 for velum opening; for glottal opening, it was 1.00 in the two lexical programs and 0.60 in the SPA sequence. These amplitudes are dimensionless values on the 0--1 reference scale defined in Section~\ref{sec:amplitude-reference}, where 1 denotes the full reference command. The revised gesture command amplitudes were 0.90 for all velum-opening gestures and 0.70 for all glottal-opening gestures. Ramp rate factors remained 0.05 for vowels and velum gestures and 0.20 for oral consonantal and glottal gestures. Thus, both changes in gesture command amplitude were evaluated together, without changing the ramp rate factors.

Each simulation lasted 2.5~s at a 1-ms integration step. Outcomes were evaluated from retained 1-ms traces filtered with a 10-ms probe synapse. A neural oscillator counted as activated if its decoded amplitude exceeded 0.3 for at least 10 consecutive samples; a gesture ramp counted as activated if its decoded activation exceeded 0.2 for at least 10 consecutive samples. Complete-sequence success required activation of every prescribed neural oscillator and ramp, strictly ordered neural oscillator starts, and strictly ordered earliest gesture starts across syllable slots. The statistical unit was a complete simulation, rather than an individual gesture. These criteria assess activation reliability, not exact movement timing, intelligibility, or mathematical stability. The revision was selected after inspecting the original failures and a separate isolated-syllable amplitude comparison. The same seeds were then reused, with the three earlier failed seeds examined first; the revised simulation subcorpus therefore supports a paired development check, not a held-out validation test. The outcome of all 120 simulation runs was independently recalculated from the stored traces, and scientific artifact checksums were verified.

%% file: 05-simulations.tex
\section{Results}
\label{sec:results}
The principal result is that a spiking neural model can generate coordinated articulatory time courses within the gestural approach (Articulatory Phonology; \citealp{browman1992,saltzman1989}). In the implemented NEF--SPA architecture, neural oscillators and neural ramp integrators provide the dynamical building blocks for syllable progression, gesture recruitment, and temporal overlap. Timing is thus realized directly through the evolving states of the spiking neural populations. Prescribed phasing rules and dynamical parameters specify the coordination pattern; recurrent neural dynamics turn these specifications into successive, overlapping gesture activations and, through the articulatory interface, vocal-tract movements.

After the initial start command, the timing and gesture network generates these activations online, step by step: an active neural oscillator recruits its gestures and initiates the next neural oscillator when the relevant phase is reached. This mechanism does not require a complete hierarchy of absolute word, syllable, and gesture time windows to be calculated before production. Syllable order and relative gesture phase windows are supplied as production knowledge, while the ongoing neural states determine their realization in time. A single online forward simulation extending through the opponent populations to articulatory movement remains the next implementation step (\ref{app:execution}). The results below document the resulting movement sequences and assess the reliability of gesture activation across neural realizations.

\subsection{Overview of word, syllable, and syllable sequence programs}
\label{sec:results-overview}
The model generated recorded articulatory realizations for all 30 selections in the input corpus, covering words, isolated syllables, and the four-syllable sequence. The nine word realizations used 9,700--14,800 neurons in their timing and gesture networks (Table~\ref{tab:words}). Across the 20 isolated-syllable selections, the corresponding counts ranged from 4,900 to 12,100. Each of these word and isolated-syllable networks used an additional 5,400 neurons for opponent control. The SPA-controlled four-syllable sequence used 309,820 neurons for timing, selection, and routing, plus 4,800 opponent neurons. These counts document the neuronal resources used to obtain the reported realizations. Within the networks with preselected syllables, each additional gesture occurrence contributes 1,200 pulse and ramp neurons, including occurrences with shared targets (Eq.~\eqref{eq:count}). For arbitrary combinations of up to four syllables from a specified inventory, we envisage one fixed network with an estimated execution core of 54,400 neurons plus syllabary, selection, and routing circuitry (Section~\ref{sec:syllable-reuse}).

The resulting neural activation trajectories and vocal-tract movements can be inspected together in the \href{https://www.speechtrainer.eu/phono2gesture-pathway/}{visual simulation player}. This synchronized visualization is a concrete output of the model: it makes the progression from neural activity to articulatory movement directly observable during playback. Table~\ref{tab:words} also reports the weakest gesture peak in each word realization; differences between rows reflect the gestures and neural realizations involved. The examples below proceed from single-syllable coordination to increasingly complex realizations.

\begin{table}[!htbp]
\centering\small
\caption{Neuron counts and gesture activation peaks in the nine recorded word realizations. $G$ is the number of separate gesture occurrences. The neuron count includes neural oscillator, next-syllable, pulse, and ramp populations; each program additionally uses 5,400 neurons in the opponent stage. The final column gives the smallest peak decoded gesture activity within the recording, rounded to two decimals. Program codes are explained in \ref{app:codes}.}
\label{tab:words}
\setlength{\tabcolsep}{5pt}
\input{words}
\end{table}
\FloatBarrier

\subsection{From a neural syllable oscillator to gestures: bap}
\label{sec:single-syllable-results}
\begin{figure}[!htbp]
\centering
\includegraphics[width=.97\linewidth]{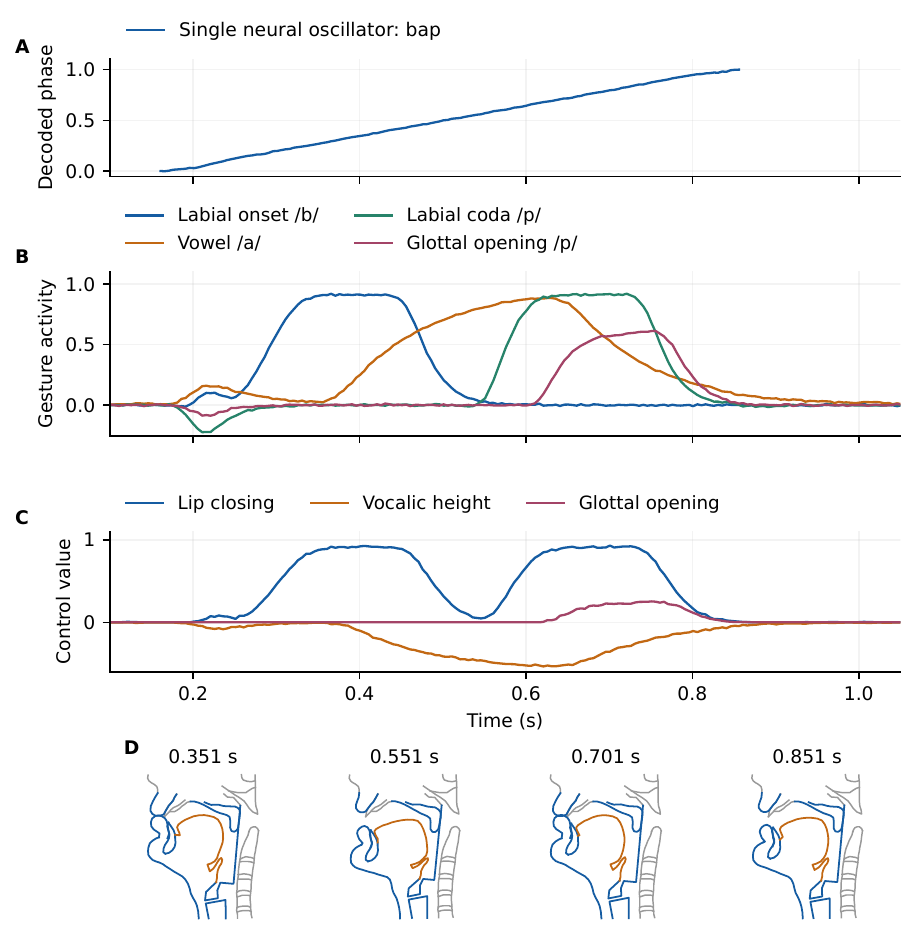}
\caption{Recorded realization of isolated \code{bap} from the original reference recordings. (A) Phase of its single neural syllable oscillator; samples with amplitude below 0.3 are omitted, and phase wraps are broken. (B) The four separate decoded gesture activations, including distinct labial onset and coda gestures. (C) Selected signed opponent outputs before geometric gains; negative vocalic-height values specify the lowering direction. (D) Stored midsagittal frames at four sample times; the tongue is highlighted in orange. All panels retain the timestamps and filtering of the same recorded cascade (Section~\ref{sec:recording}). Visualization of stored simulation outputs using AI-assisted plotting code.}
\label{fig:bap}
\end{figure}

Before examining coordination across syllables, we first establish how the gestures of one syllable unfold under a single neural oscillator. Isolated \code{bap} provides a simple CVC example: the recorded realization combines an initial labial constriction for /b/, the vowel /a/, a final labial constriction for /p/, and glottal opening for the voiceless coda. Figure~\ref{fig:bap} relates their activation to neural oscillator phase, articulatory control, and the resulting vocal-tract movement.

\FloatBarrier

\begingroup\interlinepenalty=10000
The advancing neural oscillator phase recruits the four gestures at different times (panels A and B). At the descriptive threshold of 0.2, the initial labial gesture, vowel, final labial gesture, and glottal-opening gesture first become active at 0.271, 0.396, 0.561, and 0.631~s, respectively. The vowel overlaps both labial gestures, while glottal opening overlaps the final labial gesture. The two constriction gestures retain distinct activation trajectories despite sharing the same labial target. The result is an overlapping pattern of within-syllable activity generated under one neural phase reference.\par\endgroup

The resulting lip-control trajectory contains two separate rises with a release between them, while the vocalic-height and glottal-opening parameters retain their own time courses (panel C). The stored frames show the associated movement from the initial labial constriction through the vowel to the final constriction and its release (panel D). Thus, the single-syllable realization already demonstrates the complete recorded progression from neural oscillator phase to coordinated articulatory movement. It can also be inspected in the \href{https://www.speechtrainer.eu/phono2gesture-pathway/}{visual simulation player}.

\subsection{Consonant clusters within a syllable}
\label{sec:cluster-results}
Consonant clusters increase the coordination demands within a single syllable. We therefore turn from the four gestures of \code{bap} to \emph{Schrank}, whose eight gesture occurrences are coordinated by one neural syllable oscillator. The model realized its onset and coda clusters while retaining separate gesture activations within that shared phase reference. In \emph{Schrank}, the two coda dorsal actions have the same oral target and distinct phase windows, while velum and glottal actions belong to different consonantal positions. The recorded neural activations preserve these relationships as their effects combine in the articulatory parameters. The stored trajectories thus demonstrate the neural realization of the phase specification in Figure~\ref{fig:phase}. The \emph{Stuhl} and \emph{Topf} recordings likewise show separately timed oral constrictions and associated glottal actions. The resulting cluster movements can be inspected in the \href{https://www.speechtrainer.eu/phono2gesture-pathway/}{visual simulation player}.

\subsection{From two syllables to the articulation of Kanne}
\label{sec:kanne-results}
\begin{figure}[!htbp]
\centering
\includegraphics[width=.97\linewidth]{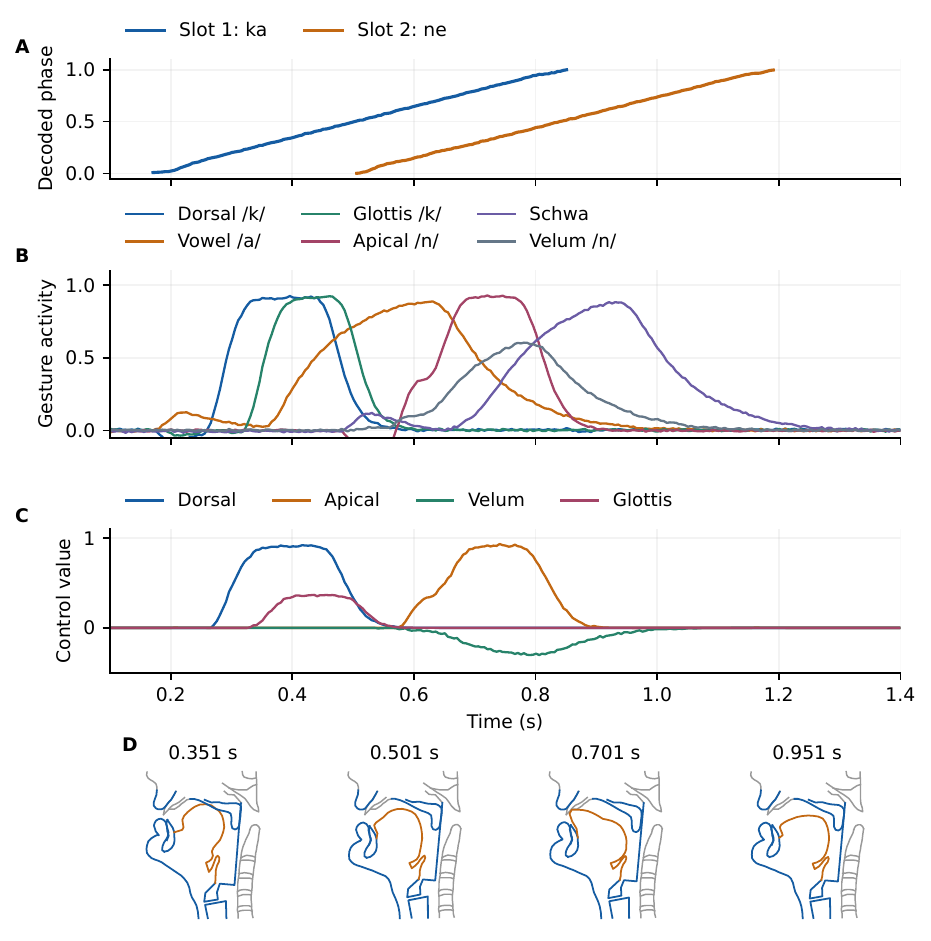}
\caption{Recorded realization of \emph{Kanne}. (A) Phase calculated from each decoded neural oscillator state; samples with amplitude below 0.3 are omitted, and lines are broken at phase wraps. (B) Separate decoded gesture activations. (C) Selected signed outputs of the opponent populations before geometric gains. Negative velum values represent the opening direction. (D) Stored midsagittal frames at four sample times; the tongue is highlighted in orange. All panels derive from the same recorded cascade. The curves retain the timestamps and filtering specified in Section~\ref{sec:recording}. Visualization of stored simulation outputs using AI-assisted plotting code.}
\label{fig:kanne}
\end{figure}

Having established coordination within a single syllable, we now examine how gestures overlap across a syllable boundary. The two-syllable word \emph{Kanne}, represented by \code{ka--ne}, makes this interaction between successive neural syllable oscillators directly measurable. The model generated an articulatory movement sequence in which gestures of the second syllable began while those of the first were still active. Figure~\ref{fig:kanne} documents this result through four views of the same recorded cascade: neural oscillator phase, gesture activation, articulatory control, and vocal-tract geometry.

\FloatBarrier

The recorded neural oscillator activity shows that the second syllable was initiated while the first neural oscillator was still active (panel A). The first neural oscillator coordinates dorsal closure, the vowel /a/, and glottal opening; the second coordinates apical closure, schwa formation, and velum opening. On the recorded initiation signals, the first threshold crossings occur at 0.156 and 0.466~s, a separation of 310~ms. This is the observed spacing of initiation activity under the settings used for this realization.

The decoded gesture activations quantify the temporal overlap between actions (panel B). The first vowel grows while the initial dorsal action is active. Later, the apical action associated with the second syllable begins while the first vowel is still decaying. At the descriptive threshold of 0.2 on both traces, this latter overlap occupies 41 retained samples, from 0.591 through 0.791~s. Under the sample-count convention of Section~\ref{sec:measurements}, this corresponds to 205~ms of overlapping gesture activity across the syllable boundary.

The resulting articulatory trajectories include dorsal constriction for the initial oral gesture, followed by apical constriction and velum lowering for the nasal in the second syllable. Panel C displays the corresponding signed opponent outputs before the geometric gains; panel D shows the tongue, jaw, lips, and velum configurations computed from those control values. A positive gesture activation expresses the strength of an action, whereas the sign of an opponent output expresses its direction on a particular parameter. For example, velum-opening activation is positive in panel B and produces a negative signed velum control value in panel C. The geometric mapping translates that value into lowering of the velum.

The \emph{Kanne} realization therefore demonstrates the complete recorded progression from neural syllable timing through overlapping gesture activations and opponent commands to articulatory movement. The synchronized recording preserves these relationships at every exported timestamp, including the intervals between the selected frames. The \href{https://www.speechtrainer.eu/phono2gesture-pathway/}{visual simulation player} makes this result accessible as playback of neural activity alongside the resulting vocal-tract movements.

\subsection{Alternative vowel realizations}
\label{sec:variants}
A further contrast concerns changing the vocalic organization while retaining the same two-syllable word structure. The two \emph{Teller} recordings produced alternative vocalic realizations under the same two-syllable organization. Both use two neural syllable oscillators, while their second syllable differs in its vowel gestures. The long-ER realization contains two overlapping vowel commands for the second neural oscillator, and the short-ER realization contains one. This difference was implemented with one additional gesture occurrence and 1,200 additional pulse and ramp neurons (Table~\ref{tab:words}). The recordings therefore demonstrate that a single neural syllable oscillator can coordinate a changing vocalic target as well as a single vowel gesture. The two alternative realizations can be compared in the \href{https://www.speechtrainer.eu/phono2gesture-pathway/}{visual simulation player}.

\subsection{Repeated syllables in the SPA-controlled sequence}
\label{sec:repeated-results}
\begin{figure}[!htbp]
\centering
\includegraphics[width=\linewidth]{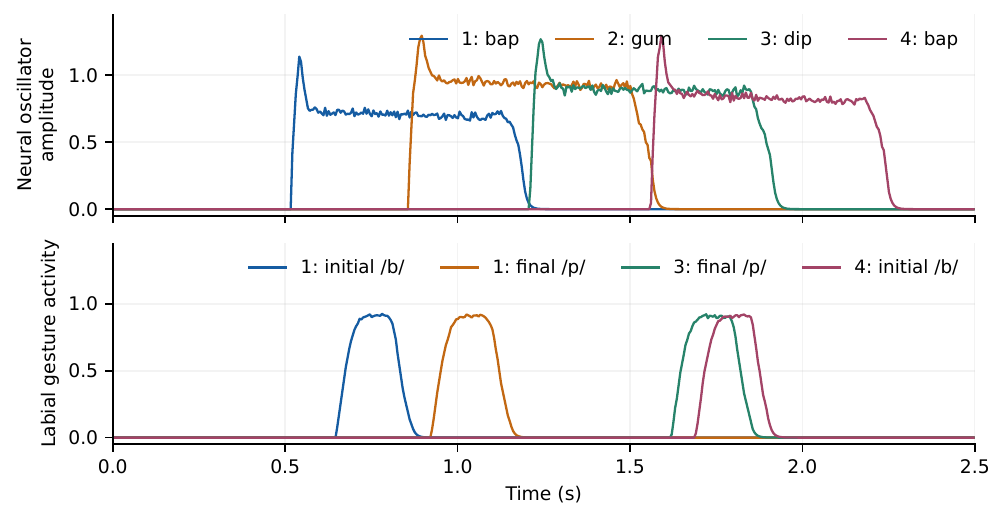}
\caption{Occurrence-specific activity in the adapted SPA-controlled sequence. The time axis includes the prescribed sequence-loading interval (\ref{app:task-settings}). Top: amplitudes of the four decoded neural oscillator states. Bottom: selected labial gesture occurrences, including the final action of \code{dip} and the initial action of the repeated \code{bap}. These curves come from separate neural ramp populations. The current adaptation retains the earlier controller while separating gesture occurrences and applying the current phasing rules. Visualization of stored simulation outputs using AI-assisted plotting code.}
\label{fig:repeat}
\end{figure}

The final individual example extends the sequence to four syllables and includes repetition of the same syllable specification. This tests whether successive occurrences retain distinct temporal states and gesture activations. The adapted SPA-controlled configuration realized \code{bap--gum--dip--bap} with four successively initiated, temporally overlapping neural oscillators and distinct gesture activations for the repeated syllable. Figure~\ref{fig:repeat} documents this result through the recorded neural oscillator amplitudes and selected labial gesture activations.

\FloatBarrier

The four initiation-signal threshold crossings occur at 0.521, 0.866, 1.216, and 1.566~s. The first and fourth occurrences use the same \code{bap} program and have distinct neural oscillator states and gesture populations. Their recorded labial activation curves remain separately identifiable. Similarly, the final labial gesture of \code{dip} and the initial labial gesture of the following \code{bap} remain separate at the ramp stage even during overlap. Their contributions meet later in the parameter mixture. The result demonstrates sequential reuse of the same syllable specification while preserving the evolving activity of each occurrence. The \href{https://www.speechtrainer.eu/phono2gesture-pathway/}{visual simulation player} displays the achieved repeated-syllable coordination and the resulting movements.
\FloatBarrier

\subsection{Reference recordings across the input corpus}
The preceding examples show how gesture coordination produces particular articulatory movement sequences. We now examine whether gesture recruitment, ordered syllable initiation, and subsequent decay of activity are also observed across the entire input corpus. Every prescribed gesture became active in each of the 30 original reference recordings, and all exported gesture traces contain finite values. Taking the maximum activation reached by each individual gesture trace, these peak amplitudes range from 0.54449 to 0.92830 (approximately 0.54--0.93). These are decoded gesture activation values on the dimensionless reference scale described in Section~\ref{sec:amplitude-reference}; the range describes the individual peaks, not all activation values over time. Initiation markers occur in the specified order in every multisyllabic recording.

To assess residual gesture activity after production of the selected utterance, we examine the last 20 stored activity values of each gesture trace at the end of its recording. Values are stored at 5-ms intervals, so these 20 samples constitute a 100-ms observation window under the sample-count convention of Section~\ref{sec:measurements}. For each gesture separately, we calculate its mean decoded activation over this final window and then take the absolute value of that mean, expressing its distance from the resting reference of zero. The largest such value across all gestures in all 30 recordings is 0.0174225: even this largest mean residual is less than 1.8\% of the reference activation level of 1. This quantity concerns activity remaining after production, whereas the peak range above concerns the strongest activation reached during production.

The decay matters because gesture activation must be transient: a gesture needs to be recruited for its articulatory contribution and then released, so that it does not continue to drive articulation after the intended movement sequence. The small final-window means indicate little sustained residual activation. Individual decoded values can still fluctuate around zero; however, the largest absolute individual value in these final windows is also only 0.02494. Thus, the small means are accompanied by small residual fluctuations, rather than merely cancellation of large positive and negative values. This is a check on the termination of gesture activation within the recorded trials; it does not require individual spiking neurons to become silent.

Taken together, the examples in Sections~\ref{sec:results-overview}--\ref{sec:repeated-results} and the corpus-wide checks reported here document complementary aspects of the result. The examples make the coordination of gestures and the resulting vocal-tract movements visible, while the checks establish that every prescribed gesture was recruited, syllables were initiated in the intended order, and gesture activity returned close to zero across the entire tested input corpus. The conclusion therefore rests on both the illustrated movement sequences and the measurements across all 30 reference recordings.

\subsection{Activation reliability with original and revised gesture command amplitudes}
\label{sec:paired-results}
Complete-sequence activation success increased from 57 of 60 runs with the original gesture command amplitudes to 60 of 60 runs with the revised amplitudes (Table~\ref{tab:paired-reliability}). All prescribed neural syllable oscillators activated in all 120 simulations. Under the original gesture command amplitudes, three realizations of the SPA-controlled sequence failed to activate the second-slot velum gesture (seeds 302, 311, and 316). The corresponding complete sequences therefore failed the activation criterion. Both lexical programs succeeded in all twenty original realizations.

\begin{table}[!htbp]
\centering\small
\caption{Successful complete sequences in the two paired simulation subcorpora of 60 runs each. All prescribed neural oscillators activated in every run in both subcorpora. All three original failures concerned the velum ramp in slot 2 of the SPA sequence. The same three seeds succeeded with the revised gesture command amplitudes.}
\label{tab:paired-reliability}
\begin{tabular}{lrr}
\toprule
Program & Original $c_g$ & Revised $c_g$ \\
\midrule
Kanne & 20/20 & 20/20 \\
Schrank & 20/20 & 20/20 \\
\code{bap--gum--dip--bap} & 17/20 & 20/20 \\
\midrule
Total & 57/60 & 60/60 \\
\bottomrule
\end{tabular}
\end{table}

\begin{figure}[!htbp]
\centering
\includegraphics[width=.95\linewidth]{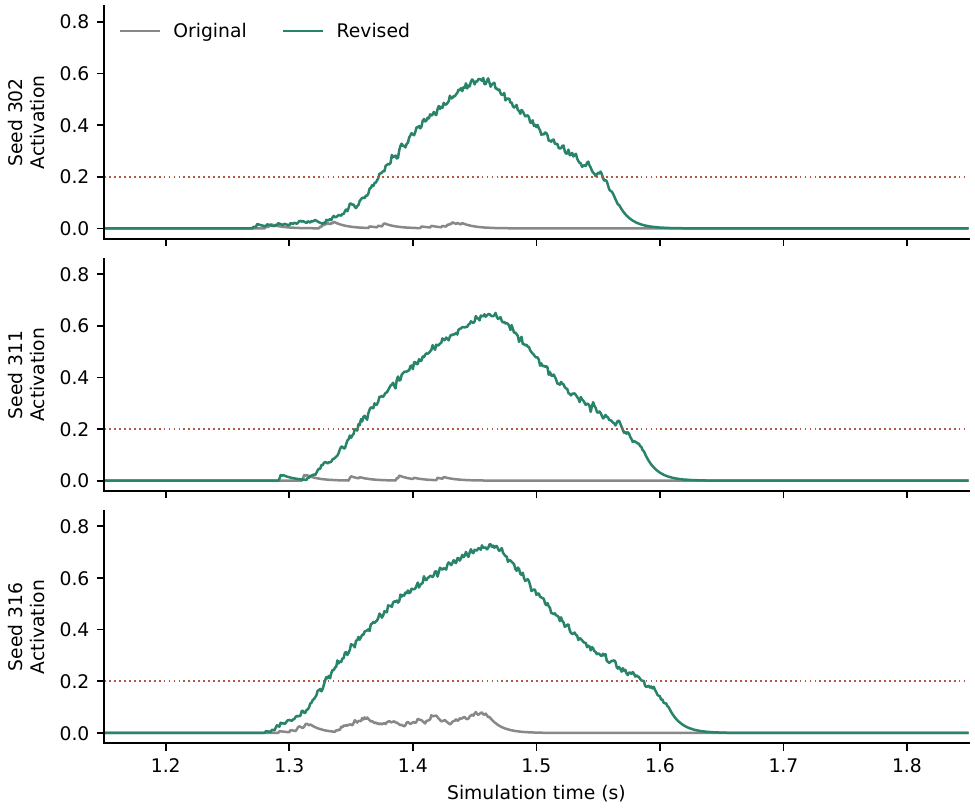}
\caption{Velum activation in slot 2 of \code{bap--gum--dip--bap} for the three originally failed neural seeds. Gray: original commands; green: revised commands, with the same neural seed and vocabulary in each pair. These are 10-ms-filtered traces retained at 1-ms spacing. The dotted line marks 0.2; the activation criterion additionally requires at least ten consecutive samples above this level. Both velum and glottal gesture command amplitudes were changed together. Visualization of stored simulation outputs using AI-assisted plotting code.}
\label{fig:paired-velum}
\end{figure}

With the revised gesture command amplitudes, all prescribed ramps and complete sequences succeeded in all 60 runs. For the three previously failed seeds (Figure~\ref{fig:paired-velum}), peak velum activations increased from 0.026, 0.022, and 0.081 to 0.583, 0.649, and 0.730, respectively. There were no newly failed realizations. Adjusting the gesture command amplitudes therefore removed the observed activation failures in the paired realizations. This result establishes successful activation under the revised settings for the tested inputs and neural seeds, according to the stated activation criterion; it does not establish unconditional stability.

\FloatBarrier

\subsection{Negative excursions in decoded gesture activation}
\label{sec:negative-results}
\begin{figure}[!htbp]
\centering
\includegraphics[width=\linewidth]{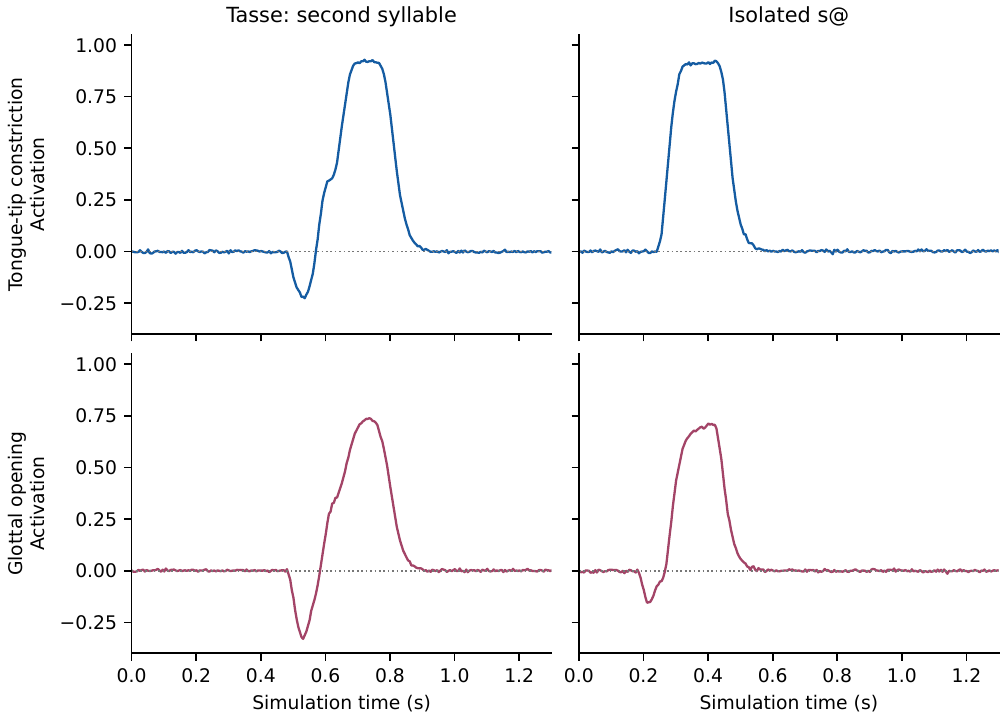}
\caption{Decoded gesture activity in the second syllable of Tasse (left) and isolated \code{se} (right), using revised gesture command amplitudes and seed 11. Columns show actual simulation time, without aligning the separate syllable onsets. The zero line distinguishes negative excursions from positive activation. Negative samples remain visible but receive zero weight at the articulatory target interface. These two illustrative recordings are separate from the two paired simulation subcorpora totaling 120 runs. Visualization of stored simulation outputs using AI-assisted plotting code.}
\label{fig:negative}
\end{figure}

The recordings with revised gesture command amplitudes revealed negative decoded excursions whose magnitudes differed between the second syllable of a word and its isolated realization. This trajectory-level variability is not captured by the binary activation criterion. In the second syllable of \emph{Tasse}, tongue-tip constriction reaches a negative minimum of $-0.227$ at 0.536~s and glottal opening reaches $-0.330$ at 0.531~s, before their main positive activations. In isolated \code{se}, the corresponding global minima are $-0.013$ and $-0.155$; the small tongue-tip minimum occurs later, at 0.786~s, rather than as a comparable preactivation dip (Figure~\ref{fig:negative}). Both examples use seed 11, but this does not make their populations or input histories identical.

The negative samples made no direct contribution to the opponent drive in either recording: replacing every negative sample by zero before the opponent-drive calculation leaves all calculated drive values exactly unchanged. This result verifies the effect of clipping at the articulatory target interface (Section~\ref{sec:articulation}) over the complete stored trajectories. It concerns the downstream target interface and does not establish that a negative state has no influence on the subsequent recurrent ramp trajectory.

\FloatBarrier

%% file: words.tex
\begin{tabular}{llrrr}
\toprule
Word / realization & Program & $G$ & Neurons & Min. peak\\
\midrule
Tasse & \code{ta} $\to$ \code{se} & 6 & 12,400 & 0.88 \\
Tasche & \code{ta} $\to$ \code{Se} & 6 & 12,400 & 0.88 \\
Kanne & \code{ka} $\to$ \code{ne} & 6 & 12,400 & 0.60 \\
Katze & \code{ka} $\to$ \code{tse} & 8 & 14,800 & 0.88 \\
Teller (short ER) & \code{tE} $\to$ \code{lA} & 5 & 11,200 & 0.88 \\
Teller (long ER) & \code{tE} $\to$ \code{lEA} & 6 & 12,400 & 0.81 \\
Topf & \code{tOpf} & 7 & 10,900 & 0.89 \\
Stuhl & \code{Stuul} & 6 & 9,700 & 0.88 \\
Schrank & \code{SraNk} & 8 & 12,100 & 0.54 \\
\bottomrule
\end{tabular}

%% file: 06-discussion.tex
\section{Discussion}
\label{sec:discussion}
The model achieves its central implementation objective: it generates coordinated articulatory movements from syllable representations using spiking neural populations in the NEF--SPA framework. Neural oscillators and neural ramp integrators organize overlapping gesture activations, whose effects pass through opponent neural control to the downstream articulatory model. The recorded realizations demonstrate this pathway for the words, syllables, and syllable sequences in the tested input corpus. This result provides the basis for discussing the architectural choices that make the realization possible, their relation to existing approaches, and the steps needed to broaden and evaluate the model.

Section~\ref{sec:syllable-reuse} examines how syllable-based neural timing could support reusable production circuits and a mental syllabary. The central issue is how stored coordination knowledge can serve changing inputs while concurrent syllable occurrences retain distinct neural states. Sections~\ref{sec:coordination}--\ref{sec:trajectory-comparison} place the architecture alongside accounts of gestural coordination and earlier neural speech-production models. These comparisons clarify what the present implementation contributes, which coordination patterns it currently takes as given, and why separating syllable timing from gesture activation dynamics creates useful modeling possibilities.

The observed activation failures motivate a closer examination of recruitment reliability (Section~\ref{sec:activation-failures}), while decoded fluctuations raise questions about the interpretation of neural variability (Section~\ref{sec:decoded-variability}). These questions matter when extending a working realization to a wider range of inputs and parameter settings. Sections~\ref{sec:integration} and~\ref{sec:evaluation} then consider the connection to lexical processing and acoustic output, the biological interpretation of the modeled populations, and comparison with human articulatory data. They identify how the demonstrated pathway can be developed into a more comprehensive and empirically tested account of speech production.

\subsection{Neural phase and reusable syllable circuits}
\label{sec:syllable-reuse}
The neural oscillator supplies a shared temporal coordinate for the gestures of a syllable. Each gesture refers to that coordinate while retaining its own onset, offset, and activation dynamics. The next-syllable command uses the same reference, linking within-syllable coordination to sequence progression. The examples, which can also be explored in the \href{https://www.speechtrainer.eu/phono2gesture-pathway/}{visual simulation player}, show how this organization supports concurrent actions and how separate occurrences retain their identities during overlap.

Three specified mechanisms contribute to the resulting timing. The recurrent transform advances neural oscillator phase, the phase windows encode relative coordination, and the ramp dynamics determine how quickly each gesture is recruited and released. These components express production knowledge through neural dynamics. Their parameters are presently assigned as part of the model specification. Learning them from experience would add an account of how that production knowledge develops.

Equation~\eqref{eq:duration} suggests a route toward speaking-rate control: changing neural oscillator frequency changes the ideal time taken to traverse a phase interval. Separating this temporal coordinate from gesture dynamics permits different responses to a rate change, as discussed in Section~\ref{sec:trajectory-comparison}. Likewise, the current chain of phase-triggered initiations provides a specific sequencing mechanism. Adaptive coupling between active neural oscillators would be an additional mechanism for stabilizing or modifying their relative phases.

Reusable syllable circuits connect the present architecture to the mental-syllabary hypothesis. \citet{levelt1994} proposed a repository linking phonological syllables to stored gestural scores, and \citet{cholin2006} reported syllable-frequency effects consistent with access to such representations. These findings motivate storage of practiced syllable-sized coordination patterns while leaving their neural implementation open. In the present architecture, the corresponding knowledge includes the required gestures and their phase windows; the neural oscillator supplies the temporal reference through which this knowledge is expressed.

A future inventory could contain circuits for individual syllables or share coordination patterns across broader syllable classes. Such classes would specify more than CV or CVC structure: for example, whether an onset contains a voiced or voiceless plosive, a fricative, or a nasal, and which consonant classes occur in onset and coda clusters. These distinctions determine the accompanying glottal-opening and velum-lowering gestures. A shared circuit would specify the phase windows for all required actions, while articulatory targets could remain selectable within that coordination pattern. This is a proposed generalization of the present syllable-specific configurations.

Reusing stored coordination knowledge must be distinguished from reusing an active neural state. A neural oscillator population could serve a later occurrence once its previous cycle has ended and a suitable initial state has been restored. If two occurrences require overlapping phase trajectories, independent neural oscillator instances are needed; one two-dimensional state cannot represent both phases independently. Gesture-activation populations also require attention because a preceding ramp can remain active after its command ends. A future implementation could allocate multiple instances of a shared circuit to concurrent occurrences and reuse them subsequently. The reported simulations instead allocate separate populations to every occurrence, including repeated syllables, so repeated reuse within one continuing network remains to be evaluated.

\begingroup\clubpenalty=10000
A fixed network for different inputs would allocate four syllable positions once and select the required coordination specifications for each utterance while retaining the same populations and connections. A capacity estimate follows from Eq.~\eqref{eq:count}: the largest single-syllable programs in the present input corpus contain eight gesture occurrences. Provisioning four neural oscillators, eight pulse--ramp pairs per position, three next-syllable command populations, and nine opponent pairs gives
\[
4\times2{,}500+4\times8\times1{,}200+3\times200+5{,}400=54{,}400
\]
neurons for this execution core. Unused positions and gesture channels would remain inactive for shorter or simpler inputs. The total network would additionally include syllabary storage, selection, routing, and SPA control; their cost depends on the supported inventory and access mechanism. This estimate therefore concerns a proposed execution capacity, not the size of a completed general-purpose model. The extension must enable neural selection of phase windows and targets within fixed connectivity and must be evaluated across different syllable combinations. Its neuron count would then remain constant across inputs within that inventory and four-position capacity.\par\endgroup

Storage at the level of individual syllables remains a candidate architecture. As a corpus-based reference, \citet{hofmann2007} report approximately 5,700 distinct phonological syllables in the German CELEX lemma database and about 11,000 in its word form database. These counts depend on database coverage and syllabification. One option in our hypothesis is an individual coordination specification for each syllable in such an inventory. The number of stored specifications is distinct from the number of neural oscillator populations required for execution: a repository could supply a smaller pool of reusable production circuits. Alternatively, a dedicated bank of 10,000 neural oscillators at the current allocation of 2,500 neurons per population would contain 25 million timing neurons, plus gesture populations, selection and routing circuitry, connections, and any concurrent copies. This is an illustrative estimate for one architecture; assessing biological feasibility requires the full storage and access mechanism. Reusing syllables across words offers a way to limit the inventory of stored motor patterns while supporting a much larger lexical repertoire.

These considerations define our working hypothesis for a mental syllabary realized with spiking neural models in the NEF-SPA framework. This sensorimotor knowledge store complements the companion mental lexicon. It contains the required gestures and the phase values delimiting their command windows. Whether entries specify individual syllables or are shared across broader syllable classes is an explicit architectural question within this hypothesis. We distinguish this relatively stable coordination knowledge from ongoing production processing: the stored specifications govern gesture recruitment as a neural oscillator advances, while its changing phase and the evolving ramp activations constitute the current execution state. This distinction is already reflected in the source code for networks with preselected syllables, where reusable phasing rules and syllable-to-gesture mappings determine the phase-dependent connections constructed for a selected input. The present implementation supplies these specifications explicitly; learning them and organizing a larger neurally accessible syllabary remain further steps.

\subsection{Coordinative structures, gesture phasing, and sensorimotor models}
\label{sec:coordination}
\citet{kelso1986} places speech within a general account of movement coordination. A coordinative structure is a functional unit of cooperating effectors organized around a task. Task-specific compensation provides evidence for this organization: a jaw perturbation during a bilabial closure recruits compensating lip activity, whereas a different constriction task recruits the appropriate tongue response. Thus, coordination concerns maintaining a functional outcome with contributions from multiple components. This within-gesture cooperation should be distinguished from the temporal coordination of different gestures.

Relative phase is a central variable in the broader coordination-dynamics account. In the bimanual phenomenon modeled by \citet{haken1985}, increasing movement frequency can induce a spontaneous transition from anti-phase to in-phase coordination. Using homologous movement cycles as the reference, these modes correspond to relative phases of approximately $180^\circ$ and $0^\circ$. The transition expresses a change in the stability of a coordinated pattern. For speech, \citet[p.~51]{kelso1986} reports a rate-induced change in lip--glottis phasing during repeated /ip/ toward the pattern for /pi/. Those measured landmark phases are not simply the $0^\circ/180^\circ$ values of the idealized bimanual example.

Articulatory Phonology treats speech as an organization of overlapping articulatory gestures \citep{browman1992}, providing a natural reference for the present separation of oral, vocalic, velopharyngeal, and glottal actions. In the 1992 account, phasing statements coordinate dynamically defined landmarks of gestures. Later coupled-oscillator models make planning dynamics explicit: each gesture has an associated oscillator, coupling establishes relative phase, and a particular oscillator phase triggers its activation \citep{nam2009}. This development connects gesture generation to the broader theory of coordination while distinguishing the earlier phasing descriptions from subsequent oscillator implementations.

Our implementation adopts a related separation of temporal organization and gesture activation, with a different allocation of state variables. One neural oscillator supplies the phase for a syllable occurrence; its gestures read prescribed windows and have separate one-dimensional ramp states. Those ramps are not gesture-specific phase oscillators, and the window positions are not relative phases that settle through mutual coupling. The current model therefore implements phase-dependent recruitment but does not demonstrate self-organized in-phase/anti-phase transitions. Adding coupling among appropriate timing states, varying rate, and testing recovery after perturbations would be needed to investigate those phenomena. Similarly, the present target mixture and geometric realization do not yet implement task-specific compensatory control. Full task dynamics \citep{saltzman1989} remains a distinct modeling choice for that level.

The two approaches consequently differ in explanatory emphasis (Table~\ref{tab:planning-comparison}). The present model is more explicit about the spiking implementation of prescribed timing and gesture recruitment. Coupled planning oscillators provide a more developed dynamical account of selected coordination patterns and aspects of syllable-structure acquisition \citep{nam2009}. Our implementation does not yet derive onset--coda asymmetries, cluster organization, or preferences in acquisition: the required coordination is supplied in its syllable programs. Neural implementation detail therefore should not be equated with greater linguistic explanatory scope.

The planning account is compatible with a neural interpretation. Indeed, \citet{nam2009} use abstract neural units and a Hebbian learning scheme in their acquisition model. The more limited neurobiological question is whether the proposed planning states and coupling relations can be identified with specific neuronal mechanisms; that identification is not established by the cited computational account. The same distinction applies to our model: a functioning spiking realization demonstrates an implementation, not that the human speech system uses that particular population code or connectivity.

The empirical identification of coupling structure also requires care. \citet{byrd2009} explicitly discuss alternative coupling graphs compatible with their observations of nasal coordination and the need for further modeling or stability data to distinguish them. \citet{muecke2020} show how changes in coupling strength and the treatment of biomechanical interactions can reconcile variable phonetic patterns with a common phonological organization. Their analysis cautions against rejecting the theory merely because a simplified implementation misses a surface pattern. Our methodological conclusion is that stronger mechanistic claims require independent constraints on both coordination dynamics and their mapping to measured movement. This requirement applies equally to our prescribed phase windows and downstream realization.

The present contribution is therefore an implementation study of supplied syllable programs, with a narrower linguistic explanatory scope than the coupled planning account. The approaches may be complementary: investigating a spiking realization of coupled planning dynamics could connect explanations of coordination patterns to explicit neural recruitment mechanisms. That integration remains future work.

\begin{table}[H]
\centering\small
\caption{Different explanatory emphases of coupled planning oscillators, particularly the account of \citet{nam2009}, and the present Spiking Neural Model. The comparison concerns the cited and current implementations, rather than limits on what either framework could implement.}
\label{tab:planning-comparison}
\setlength{\tabcolsep}{4pt}
\renewcommand{\arraystretch}{1.15}
\begin{tabular}{p{2.5cm}p{5.65cm}p{5.65cm}}
\toprule
\raggedright Question\par & \raggedright Coupled planning oscillators\par & \raggedright Spiking Neural Model\par\tabularnewline
\midrule
\raggedright How is timing organized?\par & \raggedright Relative phases settle through coupling, conditional on the specified graph, coupling functions, and parameters.\par & \raggedright Prescribed windows of syllable phase recruit separate gesture ramps.\par\tabularnewline
\raggedright What linguistic patterns are explained?\par & \raggedright Accounts for selected consonant--vowel and cluster patterns, and acquisition preferences under stated learning assumptions.\par & \raggedright Realization of supplied syllable programs; emergence and acquisition of their coordination patterns are not yet modeled.\par\tabularnewline
\raggedright How explicit is neural implementation?\par & \raggedright Dynamical planning states; the learning model includes abstract neural units and Hebbian attunement, without an explicit spiking circuit.\par & \raggedright Timing and recruitment are implemented with spiking populations and specified synaptic transformations.\par\tabularnewline
\raggedright What remains biologically unidentified?\par & \raggedright The specific neural populations and connections realizing these planning states and couplings.\par & \raggedright The human neural populations and codes corresponding to the implemented states and connections.\par\tabularnewline
\bottomrule
\end{tabular}
\end{table}

Segment labels and gesture descriptions play complementary roles in the implementation. Labels select the actions required by a syllable program; the separate gesture intervals specify how those actions overlap. For example, a nasal consonant recruits an oral constriction and velum opening whose intervals can differ. The resulting control signals describe this coordination directly. The same principle permits two successive actions on one articulator to remain identifiable until their contributions are combined.

DIVA provides a closely related perspective on sensorimotor control. As described by \citet{tourville2011}, it combines learned feedforward commands with auditory and somatosensory feedback control. GODIVA extends this framework with representations of phonological content and structure and mechanisms that release planned speech units in sequence \citep{bohland2010}. The present model focuses on how a selected syllable program unfolds through neural phase, gesture activation, and articulatory parameters. These accounts therefore suggest concrete points of comparison: the release of successive syllables, coordination within a syllable, and correction of an ongoing movement. Incorporating sensory feedback into the present architecture would allow the last of these questions to be addressed alongside the current timing mechanism.

These comparisons concern explanatory scope and computational organization. Establishing relative empirical adequacy would require common tasks, matched inputs, and comparable movement or acoustic measurements. The illustrative reference recordings across the input corpus supply a transparent starting point for such work.

\FloatBarrier
\subsection{Relation to the earlier NEF-SPA model}
\label{sec:earlier}
The earlier model introduced sequence-position neural oscillators, syllable-dependent gesture routes, and the transformation from trigger pulses to premotor activation ramps \citep{kroeger2023}. The present work develops the articulatory interface from that foundation. Its additions are separate gesture-occurrence trajectories, parameter-specific target combination, opponent neural representations, and a synchronized contour realization. The networks with preselected syllables also connect this timing mechanism to the selected words used in the companion lexical model.

The two configurations described here support complementary investigations. The networks with preselected syllables expose the temporal and articulatory consequences of supplied syllable specifications. The SPA-controlled sequence additionally retains the selection and routing mechanisms of the earlier model. Their neuron counts therefore reflect different sets of implemented functions. The adapted sequence applies the current phasing rules and separates repeated gesture occurrences through the ramp stage; it should be understood as a documented extension of the earlier implementation.

Reproducibility involves both neural and representational choices. Network seed 11 fixes neural initialization in the adapted controller, while its semantic-pointer vocabulary uses a separate random generator. The vocabulary vectors are not included in the saved export. Reconstructing that exact neural realization therefore requires this additional random state or the vectors themselves, together with the recorded software and model settings. The stored trajectories and their checksums provide the fixed reference for the illustrative results. The additional paired simulation subcorpora (Section~\ref{sec:paired-protocol}) explicitly preserve the SPA vocabulary and hold it fixed across runs with different neural seeds; generalization across vocabularies remains to be evaluated.

\subsection{Phase decoding and separate gesture dynamics}
\label{sec:trajectory-comparison}
\citet[Section~6.2, pp.~117--119]{bekolay2016} generates gestural trajectories with rhythmic dynamic movement primitives (DMPs). A multidimensional \emph{forcing function}, decoded from neural oscillator state, interpolates a manually specified gesture score as phase advances. The function maps temporal position to gesture values; it is not a second production model replacing the neural oscillator. This neural oscillator-based implementation is evaluated in his Section~7.2. The present model instead decodes phase-window commands and uses a separate recurrent population to shape each gesture occurrence's activation (Table~\ref{tab:bekolay-comparison}).

\begin{table}[H]
\centering\small
\caption{Architectural comparison with Bekolay's production implementation (2016, Section~6.2). Neural oscillator phase supplies a temporal reference in both models; the downstream construction of gesture trajectories differs.}
\label{tab:bekolay-comparison}
\setlength{\tabcolsep}{4pt}
\renewcommand{\arraystretch}{1.15}
\begin{tabular}{p{2.5cm}p{5.2cm}p{6.1cm}}
\toprule
\raggedright Aspect\par & \raggedright Bekolay (2016)\par & \raggedright Present implementation\par\tabularnewline
\midrule
\raggedright Trajectory construction\par & \raggedright Multidimensional forcing function decoded from neural oscillator phase.\par & \raggedright Phase windows issue amplitude commands; each gesture occurrence has a separate recurrent ramp population.\par\tabularnewline
\raggedright Start mechanism\par & \raggedright Default inhibition; a start-gated error population resets neural oscillator state.\par & \raggedright Input or controller drive initiates each occurrence neural oscillator; no equivalent target-error reset population.\par\tabularnewline
\raggedright Downstream output\par & \raggedright Vector gesture score converted for VocalTractLab; neural articulatory expansion left for future work.\par & \raggedright Ramps weight articulatory targets, followed by opponent neural populations and geometric movement generation.\par\tabularnewline
\bottomrule
\end{tabular}
\end{table}

This comparison concerns the implemented pathways. Bekolay explicitly leaves the full DMP point-attractor component unimplemented \citep[pp.~115, 185]{bekolay2016}; general DMP stability properties should therefore not be treated as guarantees for every stage of his neural realization or ours.

The separate ramp populations make the formation of each activation trajectory an explicit dynamical stage. A command specifies when recruitment is requested and at what level; the population's current state and recurrent dynamics determine its subsequent response. Activation can persist after command offset, allowing earlier recruitment to influence later target combinations. The gesture command amplitude $c_g$ and ramp rate factor $v_g$ offer distinct controls, although their effects are not independent: both contribute to the initial slope in Eq.~\eqref{eq:ramp-step}.

We retain this decomposition because it makes the coordination questions in Section~\ref{sec:coordination} accessible at distinct stages: temporal organization, recruitment of individual gestures, and their combined articulatory effects. Compared with decoding a complete gesture trajectory from neural oscillator state, separate activation states permit targeted changes to gesture kinetics while retaining the phase specification, or changes to timing while retaining the ramp law. This is the theoretical motivation for accepting the additional circuitry and parameter sensitivity. Reliable recruitment remains a requirement to test and improve. Both architectures use neural oscillator phase; any relative advantage for modeling coordination remains to be tested.

Table~\ref{tab:ramp-tradeoffs} summarizes the motivation for this decomposition and its costs. These are architectural consequences and modeling opportunities, not evidence of superior performance on matched tasks. Both implementations currently rely on specified gesture programs. A forcing-function approach can also represent overlapping gestures and can be modified to change individual trajectories; our contribution is to expose recruitment dynamics and their parameters in separate populations.

\begin{table}[H]
\centering\small
\caption{Advantages and costs of separate gesture-ramp populations in the present architecture. Empirical benefits require evaluation beyond the current simulation corpus recorded at a fixed nominal rate.}
\label{tab:ramp-tradeoffs}
\setlength{\tabcolsep}{4pt}
\renewcommand{\arraystretch}{1.15}
\begin{tabular}{p{2.5cm}p{5.2cm}p{6.1cm}}
\toprule
\raggedright Aspect\par & \raggedright Modeling advantage\par & \raggedright Cost or limitation\par\tabularnewline
\midrule
\raggedright Interpretable control\par & \raggedright Onset, duration, gesture command amplitude, and ramp rate factor can be adjusted for each gesture occurrence.\par & \raggedright These controls interact; gesture command amplitude also affects rise slope and finite-window peak activation.\par\tabularnewline
\raggedright Gesture state\par & \raggedright Each population retains activation history and can continue decaying after its command ends.\par & \raggedright Residual activity and transmission delays require assessment; the present response law restricts available curve shapes.\par\tabularnewline
\raggedright Rate manipulation\par & \raggedright Command timing can change while gesture response time is retained or adjusted separately.\par & \raggedright Shorter commands can reduce activation; changing phase speed alone does not uniformly compress the full trajectory.\par\tabularnewline
\raggedright Neural implementation\par & \raggedright Recruitment and dynamics can be inspected and perturbed separately for each gesture.\par & \raggedright Additional populations and pathways require reliable drive; a successful neural oscillator start does not ensure every ramp is recruited.\par\tabularnewline
\bottomrule
\end{tabular}
\end{table}

Speaking rate is a particularly useful application of this separation. For a fixed phase window, increasing neural oscillator frequency shortens the command duration $\Delta t$ (Eq.~\eqref{eq:duration}). Holding $v_g$ and $\tau_r$ fixed retains the ideal ramp time constant $\tau_r/v_g$. In the unsaturated approximation, a command of amplitude $c_g$ starting from zero reaches $c_g[1-\exp(-v_g\Delta t/\tau_r)]$ at its offset. A shorter command therefore gives a lower endpoint activation even though the response time constant is unchanged. This supplies a mechanism to investigate temporal compression accompanied by reduced target attainment, rather than requiring every activation profile to be a uniformly compressed copy.

Conversely, scaling ramp response times with command duration would more nearly preserve activation shape on a normalized time axis. Exact scaling of the full neural cascade would also require accounting for other synaptic dynamics, initiation transients, and downstream transformations. These alternatives allow us to distinguish changes in syllable timing from changes in gesture recruitment kinetics. Neither the ramp rate factor nor its decoded slope is an articulator's physical velocity; movement also depends on target combination and downstream dynamics. Consequently, the architecture motivates testing how speaking rate and movement speed covary without imposing proportional scaling in advance. It does not yet establish a better fit to human rate effects: the runs in the present simulation corpus use one nominal frequency, and rate-dependent recruitment, target attainment, overlap, and movement velocities remain to be evaluated.

\FloatBarrier
\subsection{Parameter-sensitive activation failures}
\label{sec:activation-failures}
The simulation subcorpora with original and revised gesture command amplitudes should be considered together: reporting only the revised success rate would omit the observed failures of the original parameterization. Those failures concerned recruitment of a gesture ramp while the neural syllable oscillators remained active. They establish a practical limitation of activation reliability in the tested implementation, rather than evidence for divergent or mathematically unstable neural oscillator dynamics.

They also differ from the coordination transitions discussed in Section~\ref{sec:coordination}. Such a transition is a reorganization of relative timing into another coordinated pattern. Our failed velum readouts provide no evidence for a switch between phase attractors. Keeping gesture dynamics explicit motivates investigating both reliability and possible coordination changes, but does not make a missing gesture evidence of biologically useful instability.

The results motivate the hypothesis that a small ramp rate factor combined with a low gesture command amplitude can compromise reliable gesture activation. However, the ramp rate factor was not varied, and velum and glottal gesture command amplitudes changed together. The comparison therefore does not isolate an interaction between gesture command amplitude and ramp rate factor, establish a critical gesture command amplitude, or demonstrate the same effect for vowels at low amplitudes. Contributions from the neural relays and decoder realization also remain to be separated.

No failures were observed in the revised simulation subcorpus, but twenty realizations per program do not establish a universal failure rate below 5\%. Even for a prespecified setting, zero failures in twenty independent Bernoulli trials would yield a one-sided exact 95\% upper bound of approximately 13.9\%. Here the revision was informed by the same original realizations, so that calculation is only a sample-size illustration, not a confirmatory confidence guarantee. The pooled sixty trials cover different programs and should not be treated as a single homogeneous task. Systematic parameter sweeps, circuit-level diagnostics, and larger samples are reserved for a subsequent analysis study.

The ideal ramp equation also constrains the interpretation. Below its feedback ceiling, its homogeneous dynamics have eigenvalue $-v_g/\tau_r<0$ for either ramp rate factor. Lowering $c_g$ changes the driven response, not the sign of that eigenvalue. An activation failure in the spiking implementation can therefore arise from insufficient transmitted or decoded drive without implying instability of the ideal differential equation. Routing thresholds, neuronal intercepts, and finite-population approximation are candidate contributors to investigate. Work on recurrent neural integrators likewise distinguishes ideal network construction from the sensitivity of realized dynamics to connection tuning \citep{macneil2011}; its mechanisms and tasks differ from the present gesture-recruitment problem.

Bekolay addresses his start-pulse problem with a target-directed reset: an additional population drives the neural oscillator toward its intended initial state until the start signal is released \citep[pp.~118--119]{bekolay2016}. Our observed failure instead concerns downstream recruitment: all prescribed neural oscillators start, but a required ramp can remain below the activation criterion. These are distinct vulnerable stages. The failed readout alone does not localize the cause within its driving pathway or recurrent population.

In the present SPA pathway, gated phase signals pass through three additional 60-neuron relay populations before reaching each 1,000-neuron ramp. Their intercept lower bounds are 0.30, 0.02, and 0.02; nonvocalic ramp intercepts lie between 0.02 and 1. The networks for preselected words instead use a direct 200-neuron phase-command population followed by a ramp with default intercepts. These differences make transfer and threshold effects plausible candidates for the SPA velum failures. They do not establish which stage is responsible: targeted relay and ramp probes, with one parameter changed at a time, are needed. Bekolay's neural oscillator-reset remedy cannot therefore be assumed to solve the present downstream recruitment problem.

The revised velum and glottal gesture command amplitudes were chosen to preserve transient gesture profiles while improving recruitment. This is a modeling choice supported by visual inspection and the present activation checks, not yet a quantitative fit to articulographic measurements or proof that only oral constrictions can develop plateaus. The flexibility of separate ramps therefore carries a practical constraint: gesture command amplitudes, ramp rate factors, and command durations cannot be assumed to vary freely while retaining reliable recruitment. This is especially relevant to rate manipulations, which can shorten effective drive even when amplitude is unchanged. The present data do not define a safe operating range. A factorial sweep of gesture command amplitudes and ramp rate factors, extended across command durations and accompanied by intermediate pathway probes, would distinguish reduced drive from altered response dynamics. Independent seeds and vocabulary samples would then test whether the resulting parameter choices generalize. Such analyses belong to a subsequent study.

\FloatBarrier
\subsection{Decoded variability and neural realization}
\label{sec:decoded-variability}
The distinction between repeating a linguistic action and repeating its physical realization exactly was emphasized by Georg Heike in discussions in Cologne, recalled here by the author: a speaker's attempt to repeat an utterance as faithfully as possible does not eliminate production variability. This observation motivates asking how a stable linguistic outcome can coexist with variable trajectories.

An empirical illustration is provided by \citet{mefferd2019}, who analyzed jaw and tongue kinematics in 17 speakers producing ten repetitions of the same sentence in each of four speaking conditions. Endpoint variability was measurable, and its relation to speaking condition differed between articulators. Such findings support treating token-to-token variability as a property to characterize rather than assuming identical repetition. They do not establish that our decoded neural fluctuations have the same source or distribution.

The negative excursions in Section~\ref{sec:negative-results} concern decoded population variables, not negative firing rates. Their difference between isolated and embedded syllables suggests examining neural realization and input context separately. Preferred directions, intercepts, and decoder weights depend on network construction; recent input and ongoing filtered activity determine the network state when a gesture is recruited. The present examples do not isolate the contribution of any one of these factors.

The comparison must also distinguish ordinary variation in a successfully realized gesture from failure to recruit a required gesture at all. Biological variability does not by itself justify accepting missing model gestures as normal speech errors. The runs in each simulation subcorpus vary neural construction; they do not constitute a repeated-utterance experiment within one unchanged network. A closer counterpart to Heike's observation would repeat an utterance in a fixed network while explicitly controlling initial states, preceding activity, and any added stochastic input.

Spiking dynamics alone do not imply ongoing stochastic noise. With fixed construction, initial conditions, and input, the implemented simulation can produce reproducible trajectories. Changing the seed samples another realization, while changing sequence context may alter input history as well as construction. The multi-seed evaluation asks whether the same linguistic function survives these changes in neural realization. That is a useful test of robustness within a spiking implementation, but it does not identify the fluctuations with the mechanisms of variability in the human brain.

\subsection{Integrating lexical processing, articulation, and acoustics}
\label{sec:integration}
The companion lexical manuscript describes the cognitive model part, from concepts through lemmas to phonological forms. This paper describes the sensorimotor model part, from ordered syllable input through neural gesture activity to opponent neural control, and its connection to the articulatory model. Shared words and a checked syllable readout establish the cognitive-to-sensorimotor interface; decoded parameter trajectories establish the sensorimotor-to-articulatory interface. Continuous execution across this pathway would join these existing components more closely.

The next integration steps are concrete. An accepted lexical output can initiate the corresponding syllable program automatically, including its selected realization variant. Extending the syllable readout to further words, and implementing its ordered output neurally, would broaden the lexical interface. Sensory feedback could then influence timing or articulatory control during the ongoing production process. At present, the reported cascade proceeds from the supplied program to movement without corrective feedback from its output.

The NEF-SPA approach provides a framework for linking these cognitive and sensorimotor components. SPA supplies structured representations and selection mechanisms, while the NEF supplies methods for continuous representation, transformation, and dynamical control. Spaun illustrates this combination in another domain: it receives visual images and produces its responses with a physically modeled arm \citep{eliasmith2012}. This example demonstrates how cognitive processing can be connected to sensory input and motor output within the same framework. For speech, the analogous objective is a pathway that reaches articulatory movement and sound and can use the resulting sensory information.

The acoustic extension of DYNARTmo already provides a separate downstream component relating articulatory geometry, aerodynamics, and sound generation \citep{kroegerAcoustics2026}. The present exporter supplies continuous control trajectories and gesture-specific constriction labels as a starting point for that connection. Integration requires compatible treatment of full, fricative, and lateral constrictions, nasal pathways, phonation, and aerodynamic drive. For example, preserving a lateral-constriction label identifies an intended action, while acoustic realization also requires the corresponding air passage. The recordings presented here end at geometric articulation; they do not yet include an acoustic waveform. A joined simulation would make both the movement and the generated sound available for evaluation.

\subsection{Biological interpretation and further evaluation}
\label{sec:evaluation}
The sensorimotor model part describes a possible neural realization of temporal coordination and articulatory control. Functionally, the neural oscillators and gesture ramps provide premotor timing and recruitment, while the opponent populations form the output stage toward neuromuscular control. The implementation represents this output with abstract opposing parameter channels that preserve concurrent commands. Relating these channels to anatomical muscles, muscle forces, or mechanical stiffness requires a more detailed physiological model. Connecting the neural timing and control functions to particular human neuronal populations remains an empirical task. The downstream articulatory model converts the decoded parameter trajectories into movement kinematically; tissue mechanics would extend this peripheral realization.

The simulation corpus serves the paper's primary aim of making this architecture transparent. It shows the behavior of selected programs under documented configurations and supports inspection of sequencing, overlap, and the resulting contours. These recordings do not establish intelligibility, agreement with human articulatory trajectories, or general robustness across programs, vocabularies, and parameter settings. The paired simulation subcorpora provide a limited activation-reliability check; broader validation requires further simulations and corresponding empirical data.

German syllable cut offers a specific test of vowel--consonant coordination. Electromagnetic articulography analyses by \citet{hoole2002} indicate tighter coupling between opening and closing movements for lax vowels, while also showing that their movement patterns cannot generally be derived by simply truncating those of tense vowels. Acoustic and articulatory results from \citet{belz2023} further show that lax vowels can lengthen phrase-finally, highlighting the role of prosodic context. For the medial consonants assigned to the second syllable here, adjusting the initiation time of that syllable's neural oscillator offers a direct way to change their overlap with the preceding vowel. This timing adjustment provides a candidate mechanism for investigating syllable cut while preserving a single realization of the medial consonant. The present simulations keep the phase window for intersyllabic initiation fixed; whether varying it adequately captures the observed contrast remains a question for comparison with human articulatory data.

Several extensions follow directly from the present description. Further neural seeds can extend the present activation-reliability checks while holding the program fixed. Changes in phase windows, neural oscillator frequency, and population size can reveal which aspects of sequencing and overlap depend on particular parameter choices. Additional words, contexts, and stress patterns can test a wider range of gestural coordination and prosody. Comparisons with human articulatory recordings should distinguish command boundaries, neural activation landmarks, and geometric movement. Acoustic integration would further allow perceptual assessment of the same programs. These evaluations can build on the current model description and recorded examples in subsequent studies.

%% file: 07-resources.tex
\section{Code, data, and reproducibility}
\label{sec:resources}
The model code and recorded examples are organized in the public \href{https://github.com/bjkroeger/nef-spa-phono2gesture}{Phonology-to-Gesture repository}. The companion lexical resource and retrieval implementation are provided in the \href{https://github.com/bjkroeger/nef-spa-mental-lexicon}{Mental Lexicon repository}. The \href{https://www.speechtrainer.eu/phono2gesture-pathway/}{visual simulation player} and \href{https://www.speechtrainer.eu/nef-spa-phono2gesture-pathway.htm}{explanatory documentation} are available on the author's SpeechTrainer website.

The public player provides playback of recorded activity and synchronized articulation. Local execution provides the separate facility for running supported neural simulations. For each selection, the exported files retain neural oscillator states, gesture activity, opponent drives and outputs, articulatory parameters, and contour frames. Metadata identify the relevant timing version, neural counts, seeds, and source hashes. The publication snapshot additionally includes a checksum manifest for the distributed assets.

The manuscript's figure-generation script reads these recorded outputs directly and writes the figures, word-program table, and machine-readable measurements. Its provenance file lists SHA-256 checksums for the recordings and relevant source files used for the reported figures. Recreating the figures from the stored outputs is distinct from regenerating those outputs with a simulator. Re-simulation additionally depends on software versions, random-state handling, and the model configuration, including the vocabulary state of the adapted SPA-controlled sequence discussed in Section~\ref{sec:earlier}. The respective repository notices specify the reuse terms for each component.

The two paired simulation subcorpora and their verification records are publicly available in a versioned data release in the Phonology-to-Gesture repository \citep{kroegerData2026}. The second simulation subcorpus combines the completed twenty SPA runs and forty lexical runs, with matching model-source hashes. The release includes raw traces, vocabulary vectors for the SPA runs, protocol files, per-run outcomes, checksums, and a paired 60-row comparison. It also retains the thirty historical reference recordings and the two additional negative-excursion examples. The original reference figures remain tied to their archived recordings; later web-player recordings can use a different parameterization or neural realization.

%% file: 08-conclusion.tex
\section{Conclusion}
This paper demonstrates that a sensorimotor model built with spiking neural populations in the NEF--SPA framework can realize the pathway from phonological syllable representations to coordinated articulatory movements. Temporal progression arises directly from the evolving neural dynamics under the specified phasing rules. Within the sensorimotor model part, neural syllable oscillators provide a temporal reference for relative gestural organization. Phase-dependent commands coordinate overlapping actions, recurrent populations generate their activation trajectories, and opponent neural outputs supply articulatory control values. The downstream articulatory model converts these parameter trajectories into vocal-tract movement, making the transition from neural control to peripheral realization directly inspectable.

The illustrative reference recordings document this progression across the input corpus of selected lexical words, isolated syllables, and a sequence with a repeated syllable. The recorded trajectories make timing, target contributions, and articulatory movement available for inspection; the \href{https://www.speechtrainer.eu/phono2gesture-pathway/}{visual simulation player} presents these relationships through synchronized playback. Together with the companion mental-lexicon account, the present model description specifies complementary components of a production architecture extending from concepts to articulation. The explicit syllable interface, recorded neural trajectories, and articulatory parameter exports provide a foundation for integrated simulation and for the subsequent connection to acoustics.

The paired simulation subcorpora add a practical qualification to this architecture: prescribed temporal organization does not guarantee reliable gesture recruitment in every neural realization. The original settings produced three velum-activation failures in the SPA sequence; adjusted gesture command amplitudes eliminated these failures in the same twenty seeds without introducing failures in the forty lexical runs. Reporting both subcorpora makes this parameter sensitivity explicit. Larger independent samples and a systematic analysis of the circuit remain necessary before making general claims about stability.

%% file: declarations.tex
\begin{samepage}
\section*{CRediT authorship contribution statement}
\textbf{Bernd J. Kröger:} Conceptualization, Methodology, Software, Formal analysis, Investigation, Data curation, Visualization, Validation, Writing -- original draft, Writing -- review and editing.

\end{samepage}
\section*{Funding}
This research did not receive any specific grant from funding agencies in the public, commercial, or not-for-profit sectors.

\section*{Declaration of competing interests}
The author declares that he has no known competing financial interests or personal relationships that could have appeared to influence the work reported in this paper.

\section*{Data availability}
The frozen simulation data, source snapshots, and verification records are publicly available in a versioned GitHub data release \citep{kroegerData2026}. Model code and reference examples are also accessible through the repositories and visual simulation player listed in Section~\ref{sec:resources}.

\section*{Declaration of generative AI and AI-assisted technologies in the manuscript preparation process}
During the preparation of this work, the author used ChatGPT and Codex (OpenAI) to assist with drafting and revising manuscript text, programming, and preparation of analysis and visualization scripts. Figures were generated programmatically from the specified model definitions or stored simulation outputs. The author reviewed and edited the resulting material and takes full responsibility for the content of the publication.

%% file: appendix.tex
\section{Implementation parameters and notation}
\label{app:parameters}
\subsection{Neural configuration}
Table~\ref{tab:neural-parameters} collects the settings of the neural architecture. The networks with preselected syllables instantiate the required gesture paths directly; the adapted SPA-controlled sequence additionally retains the earlier controller and routing populations.
\begin{table}[H]
\centering\small
\caption{Principal neural parameters for networks with preselected syllables. Intercepts specify where neurons begin responding along their preferred direction. Seeds fix sampled neural parameters for a given construction; vocabulary vectors require their own specification (Section~\ref{sec:earlier}).}
\label{tab:neural-parameters}
\begin{tabular}{ll}
\toprule
Quantity & Value\\
\midrule
Neuron model & LIF\\
Neural oscillator & 2 dimensions; 2,500 neurons per occurrence\\
Neural oscillator intercept distribution & Uniform on $[0.3,1]$\\
Nominal frequency; recurrent synapse & 1.5 Hz; 25 ms\\
Neural oscillator amplitude threshold & Neural oscillator norm at least 0.3\\
Gesture pulse / next-syllable population & 1 dimension; 200 neurons\\
Pulse / next-syllable intercept distribution & Uniform on $[0.02,1]$\\
Gesture ramp & 1 dimension; 1,000 neurons\\
Ramp rate factor $v_g$ & 0.05 (vowel, velum); 0.20 (other)\\
Ramp feedback function & $\min(0.9,r_g)$\\
Default connection synapse & 5 ms, where not otherwise specified\\
Opponent channel & 1 dimension; 300 neurons\\
Opponent input connection & No synaptic filter\\
Opponent encoders; intercepts & Positive; uniform on $[0,1]$\\
Illustrative network / opponent seed & 11 / 23\\
\bottomrule
\end{tabular}
\end{table}

\subsection{Task and recording settings}
\label{app:task-settings}
Table~\ref{tab:recording-parameters} summarizes the procedure introduced in Section~\ref{sec:task}. These settings determine how a program is initiated and observed with the specified architecture. In the SPA-controlled runs, the four syllable--position associations are loaded during 0--0.4~s, followed by the external start command during 0.4--0.45~s. These are prescribed input intervals on the simulation clock. They are not estimates of biological utterance-preparation time; the minimum loading duration required by the implemented network has not been evaluated.
\begin{table}[H]
\centering\small
\caption{Input, simulation, and observation settings. Probe filters define recorded estimates. Filtered gesture recordings also supply the input to the subsequent opponent stage.}
\label{tab:recording-parameters}
\begin{tabular}{ll}
\toprule
Quantity & Value\\
\midrule
Input pulse (preselected syllables) & 0.15--0.20 s; unit amplitude\\
SPA sequence loading (prescribed) & Four 100-ms syllable-position inputs\\
SPA start command & 0.40--0.45 s\\
Timing-network simulation duration & 2.5 s\\
Integration step & 1 ms\\
Neural oscillator, ramp, initiation probe filters & 10 ms\\
Opponent probe filter & 10 ms\\
Playback recording grid & 500 samples; 0.001--2.496 s; 5-ms spacing\\
Paired simulation subcorpora: raw grid & 1-ms spacing; 2,500 samples\\
Illustrative initiation-marker threshold & 0.4\\
Gesture-overlap threshold & 0.2 on both selected traces\\
Settling observation window & Final 20 retained samples (100 ms)\\
\bottomrule
\end{tabular}
\end{table}

\subsection{Program codes}
\label{app:codes}
The program identifiers use an ASCII-oriented notation inherited from the lexical and articulatory resources. Upper-case \code{S} denotes the postalveolar fricative, \code{N} the velar nasal, \code{E} the open-mid front vowel, and \code{O} the open-mid back rounded vowel. Lower-case \code{e} in \code{se}, \code{Se}, \code{ne}, and \code{tse} denotes schwa. The suffix \code{A} in \code{lA} selects the short vocalic-ER target, while \code{lEA} selects the two-target variant. The \code{uu} in \code{Stuul} is part of the syllable identifier; the program contains one vocalic gesture at the /u/ target. These codes name implemented programs and are not a complete phonetic transcription system.

\subsection{Source organization}
The networks for preselected words and the explicit realization mapping are defined in \path{lexical_model.py}. Shared phase windows and phasing rules are in \path{timing.py}. The SPA-controlled sequence is adapted by \path{simulate.py} and \path{sequence_gestures.py}, retaining the controller in \path{source/original_model.py}. The target mixture and opponent replay are implemented in \path{muscles.py}; these populations form the output stage of the sensorimotor model part and represent abstract parameter channels toward neuromuscular control. Geometric conversion and parameter export are implemented in \path{sagittal.py}. The companion lexical software implements the graph readout in \path{syllabification.py}.

The figure-generation script reads the stored traces, computes the descriptive measurements defined in Section~\ref{sec:measurements}, and generates the plots. The associated checks verify signal presence, finite values, initiation order, and settling under the stated criteria. Evaluation of phonetic accuracy requires the additional comparisons discussed in Section~\ref{sec:evaluation}.

\subsection{Simulation execution and planned online integration}
\label{app:execution}
The reported simulations use two computational stages. First, the neural oscillator and gesture network is simulated over the complete trial interval. Filtered gesture activation trajectories are retained on the 5-ms recording grid. The numerical mapping in Section~\ref{sec:articulation} then converts these samples into positive and negative commands for the opponent neural populations. A second neural simulation replays these commands, using linear interpolation onto its 1-ms integration grid. Both members of every opponent pair are simulated together in this second stage. Their outputs are sampled at the original recording times and passed to the geometric adapter. Thus, all displayed signals share the same model-time coordinates, although the two neural stages are computed in separate runs.

This execution procedure is a technical limitation of the present implementation relative to our intended model of ongoing speech production. The intended architecture advances from syllable timing through gesture activity and opponent control to articulatory movement as the utterance unfolds. It does not require completion of an entire utterance before the opponent populations can be updated. The current replay procedure is documented here for reproducibility; the reported results were obtained with this procedure.

The next model version is planned to replace whole-trial replay with a single forward simulation. At each numerical integration step, the available gesture readouts would determine the two opponent inputs, and both positive and negative populations would advance on the same time grid. The existing 1-ms integration step can be retained independently of the 5-ms interval used for recording and display. If computation is organized in 5-ms blocks, each block would contain five successive 1-ms integration steps. This arrangement would propagate activity as it becomes available, without waiting for the complete utterance or interpolating from future stored samples. Any retained readout filters would be applied causally during the ongoing calculation. The integrated implementation will require new simulations and comparison with the present results because the current intermediate sampling, interpolation, and rounding can affect the trajectories.

Here, \emph{online} means that all stages advance during ongoing production; it does not require the numerical simulation to run at real-time speed on the computer. Combined with the incrementally renewed syllable preparation window described in Section~\ref{sec:task}, the planned implementation is intended to support utterances extending over many seconds or minutes. This also requires allocation and reuse of active neural circuits and verification of stable long-duration operation. Continuous production on that scale has not been demonstrated by the present short-input simulations.